\documentclass[sigconf]{acmart}

\AtBeginDocument{%
  }

\copyrightyear{2025} 
\acmYear{2025} 
\setcopyright{acmlicensed}
\acmConference[KDD '25]{Proceedings of the 31st ACM SIGKDD Conference on Knowledge Discovery and Data Mining V.1}{August 3--7, 2025}{Toronto, ON, Canada} \acmBooktitle{Proceedings of the 31st ACM SIGKDD Conference on Knowledge Discovery and Data Mining V.1 (KDD '25), August 3--7, 2025, Toronto, ON, Canada} \acmDOI{10.1145/3690624.3709343} 
\acmISBN{979-8-4007-1245-6/25/08}

\usepackage{soul}
\usepackage{url}
\usepackage[utf8]{inputenc}
\usepackage{amsthm}
\usepackage{algorithm}
\usepackage{algorithmic}

\graphicspath{ {./} }
\usepackage{multirow}

\begin{document}

\title{An Efficient Diffusion-based Non-Autoregressive Solver for Traveling Salesman Problem}


\author{Mingzhao Wang}
\affiliation{
  \institution{Jilin University}
  \city{Changchun}
  \country{China}}
\email{wangmz22@mails.jlu.edu.cn}

\author{You Zhou}
\authornote{Corresponding Author.}
\affiliation{
  \institution{Jilin University}
  \city{Changchun}
  \country{China}}
\email{zyou@jlu.edu.cn}

\author{Zhiguang Cao}
\affiliation{
  \institution{Singapore Management University}
  \country{Singapore}}
\email{zhiguangcao@outlook.com}

\author{Yubin Xiao}
\affiliation{
  \institution{Jilin University}
  \city{Changchun}
  \country{China}}
\email{xiaoyb21@mails.jlu.edu.cn}

\author{Xuan Wu}
\affiliation{
  \institution{Jilin University}
  \city{Changchun}
  \country{China}}
\email{wuuu22@mails.jlu.edu.cn}

\author{Wei Pang}
\affiliation{
  \institution{Heriot-Watt University}
  \city{Edinburgh}
  \country{United Kindom}}
\email{w.pang@hw.ac.uk}

\author{Yuan Jiang}
\authornotemark[1]
\affiliation{
  \institution{Nanyang Technological University}
  \country{Singapore}}
\email{yuan005@e.ntu.edu.sg}

\author{Hui Yang}
\authornotemark[1]
\affiliation{
  \institution{Jilin University}
  \city{Changchun}
  \country{China}}
\email{yanghui2012@jlu.edu.cn}

\author{Peng Zhao}
\affiliation{
  \institution{Jilin University}
  \city{Changchun}
  \country{China}}
\email{pengzhao23@mails.jlu.edu.cn}

\author{Yuanshu Li}
\affiliation{
  \institution{Jilin University}
  \city{Changchun}
  \country{China}}
\email{lys21@mails.jlu.edu.cn}

\renewcommand{\shortauthors}{Mingzhao Wang et al.}

\begin{abstract}
Recent advances in neural models have shown considerable promise in solving Traveling Salesman Problems (TSPs) without relying on much hand-crafted engineering. However, while non-autoregressive (NAR) approaches benefit from faster inference through parallelism, they typically deliver solutions of inferior quality compared to autoregressive ones. To enhance the solution quality while maintaining fast inference, we propose DEITSP, a diffusion model with efficient iterations tailored for TSP that operates in a NAR manner. Firstly, we introduce a one-step diffusion model that integrates the controlled discrete noise addition process with self-consistency enhancement, enabling optimal solution prediction through simultaneous denoising of multiple solutions. Secondly, we design a dual-modality graph transformer to bolster the extraction and fusion of features from node and edge modalities, while further accelerating the inference with fewer layers. Thirdly, we develop an efficient iterative strategy that alternates between adding and removing noise to improve exploration compared to previous diffusion methods. Additionally, we devise a scheduling framework to progressively refine the solution space by adjusting noise levels, facilitating a smooth search for optimal solutions. Extensive experiments on real-world and large-scale TSP instances demonstrate that DEITSP performs favorably against existing neural approaches in terms of solution quality, inference latency, and generalization ability. Our code is available at \href{https://github.com/DEITSP/DEITSP}{https://github.com/DEITSP/DEITSP}.

\end{abstract}




\maketitle

\section{Introduction}
The Traveling Salesman Problem (TSP) is a well-known combinatorial optimization problem that formulated on a fully connected graph with non-negative edge weights \cite{Garey1990}. A TSP solution manifests as a Hamiltonian cycle, where the objective is to minimize the sum of edge weights along the tour. In this context, a Hamiltonian cycle refers to a tour that starts and ends at the same node while visiting each remaining node exactly once. TSP has diverse applications in practical scenarios, such as transportation and facility location \cite{Christiansen2004, wu_neural_2023}. Given its significance in both theoretical and practical realms, numerous exact \cite{Bellman1962, Padberg1991, Applegate2007}, approximate \cite{Bansal2004, kizilatecs2013nearest, wu_incorporating_2023}, and heuristic algorithms \cite{Helsgaun2017, Zhong2018, Onizawa2022fast} have emerged over the years. However, most of these algorithms are intricate, characterized by many manually formulated rules that heavily rely on expert knowledge.

In recent years, neural network (NN)-based models have emerged as a promising alternative for addressing the complexities of the TSP \cite{wu2024}. Although these models lack well-established theoretical guarantees, empirical evidence suggests they can achieve near-optimal solutions. NN-based models for TSP can be broadly categorized into two types based on their decoding strategies: Autoregressive (AR) and Non-Autoregressive (NAR). AR models, inspired by neural machine translation, sequentially generate nodes on the Hamiltonian cycle and have shown satisfactory performance \cite{Kool2019, Bresson2021, Kim2021, Xin2021, Gao2023, Jung2023}. However, their inherent sequential nature limits the inference speed of these networks. In contrast, NAR models generate solutions in a one-shot manner for fast inference. 

While earlier NAR models often yield lower-quality solutions, recent advancements in diffusion-based learning for TSP have shown improvements in solution quality. \citet{graikos2022} transformed TSP instances into low-resolution grayscale images using an image-based diffusion model. \citet{Sun2023} developed a graph-based diffusion model to explicitly represent problems as graph. \citet{ma2024} proposed a semi-supervised training strategy for diffusion models, while \citet{li2024} incorporated objective optimization guidance into the denoising steps through gradient feedback, enhancing sampling efficiency. Although these approaches improve performance, they often rely on simulating a Markov chain with numerous steps, which increases time overhead. Moreover, these methods tend to focus excessively on obtaining a single high-quality solution, often neglecting the exploration of the solution space, resulting in high computational costs for marginal improvements, especially when faced with test problems that differ in size and distribution from those seen during training.

To enhance generalization ability and inference speed, we propose an efficient diffusion model named DEITSP. This model directly maps noise to the optimal solution, eliminating the need for multi-step Markov processes. Unlike previous NAR approaches that generate only a single solution, DEITSP simultaneously explores multiple solutions. We introduce an efficient iterative strategy that alternates between adding and removing noise for generating multiple solutions to improve exploration and solution quality. Our approach provides flexibility by allowing users to adjust the number of iteration steps according to practical needs, balancing the trade-off between solution quality and inference speed. Specifically, DEITSP formulates the generation of adjacency matrices as edge classification tasks. Our model consists of three key components: (1) a one-step diffusion model based on a controlled discrete noise addition process and self-consistency enhancement, enabling high-quality solution prediction through single-step denoising; (2) a dual-modality graph transformer that enhances the extraction and integration of features from different modalities with fewer layers, achieving faster inference and stronger feature representation; and (3) an efficient iterative strategy that alternates between adding and removing noise to enhance exploration and refine the solution space of TSP, along with a corresponding scheduling framework that adjusts noise levels to facilitate a smooth search for solutions.

We conducted extensive experiments comparing DEITSP with 16 NN-based baselines on TSP instances ranging in size from 20 to 1000 nodes. DEITSP achieved the highest solution quality and demonstrated superior generalization, effectively scaling from small to large TSP instances. It also generalized well to instances beyond the training distribution, including real-world cases from the USA, Japan, Burma, and TSPLIB \cite{reinelt1991tsplib, Yong2024}. Ablation experiments further confirmed the effectiveness of our proposed components.

We summarize our contributions as follows:

\begin{itemize}
\item We introduce a one-step diffusion model that enables optimal TSP solution prediction through simultaneous denoising of multiple solutions.
\item We design a dual-modality graph transformer that significantly enhances feature extraction and fusion from node and edge modalities, resulting in faster inference and more robust feature representation.
\item We develop an efficient iterative strategy specifically tailored for TSP, alternating between adding and removing noise to enhance exploration and progressively refine the solution space, thereby improving solution quality.
\item We devise a scheduling framework to adjust noise levels automatically, facilitating a smooth optimal solution search.
\item We validate DEITSP through extensive experiments, demonstrating its excellence in solution quality, inference speed, and generalization across various real-world TSP instances and large-scale benchmarks.
\end{itemize}

\section{Related Work}

In this section, we review NN-based TSP solvers, which are typically categorized as either AR or NAR based on decoding methods. We also introduce discrete diffusion models relevant to our work.

\subsection{Neural Network-based TSP Solvers}

With recent advancements in deep learning, numerous NN-based models have been proposed to solve TSPs. These models can be broadly categorized into AR and NAR based on their decoding methods. It is important to note that we focus on end-to-end learning methods for solving TSPs, excluding neural improvement algorithms that combine deep learning with heuristic search, as they typically result in longer inference latency.

We first review pioneering NN-based approaches for solving TSPs using AR decoding. \citet{Vinyals2015} proposed a sequence-to-sequence supervised learning (SL) model called Pointer Network (PtrNet), which uses the Long Short-Term Memory (LSTM) architecture from natural language processing \cite{Bahdanau2014}. PtrNet inputs node coordinates, employs an attention mechanism, and generates TSP solutions via supervised learning step-by-step. While PtrNet provides a novel perspective for solving Combinatorial Optimization (CO) problems, its unsatisfactory performance makes it challenging to be applied in practical settings \cite{Li2021,Fu2021}. Subsequently, \citet{Bello2017} expanded PtrNet with an actor-critic reinforcement learning (RL) algorithm, eliminating the need for optimal solutions during training. \citet{Nazari2018} further enhanced PtrNet’s performance with node embeddings from attention layer. Recently, the Transformer has set new performance records in various applications \cite{Vaswani2017}, leading to several Transformer-based TSP models, notably \cite{Kool2019, Kwon2020, Bresson2021, Deudon2018}.  \citet{kim2022} improved generalization by leveraging symmetries like rotational and reflectional invariance. \citet{Jung2023} introduced a lightweight CNN-Transformer model using a CNN embedding layer and partial self-attention. \citet{Gao2023, jiang2023ensemble, xiao2024improving} integrated extra policies to enhance generalization. These models share the same Transformer-encoder architecture but differ in the decoder. Specifically, \citet{Deudon2018} used node information from the last three steps, \citet{Kool2019} used information from the first and last steps, \citet{jiang23a} employ contrastive learning to enhance the embedding of nodes, and \citet{Bresson2021} fused all previously output node information. All of these models demonstrate excellent performance.

However, the constructive methods used by AR models naturally disadvantage inference speed compared to NAR approaches \cite{Ran2021, Sun2023}. To address this issue, several NAR models have been developed to improve inference speed when solving TSPs \cite{Nowak2018,Joshi2019}. These models treat TSP as a link prediction task, which has been studied in various research fields \cite{Wang2019,Xiao2020,Xiao2023}. They use supervised learning (SL) to train their models to estimate the importance of each edge in the optimal solution. For instance, \citet{Nowak2018, Joshi2019} utilized a graph network based TSP solver trained with SL, which takes a TSP instance as a graph input and directly outputs a heat map representing the importance of edges. It is worth noting that during training, these models relax the sequential constraint and obtain the edge’s importance prediction by minimizing the binary cross-entropy loss between the adjacency matrix of the optimal solution and their heat map. During inference, these models apply greedy search or beam search on the heat map to obtain a feasible TSP solution. Despite the significant improvement in inference speed offered by NAR approaches \cite{Nowak2018,Xiao2023,xiao2024}, the solution quality of these models is often suboptimal. Existing NAR models generally fail to outperform most of the aforementioned AR models \cite{Joshi2019, Joshi2022, Xiao2023}. Thus, there is a pressing need to further improve existing models to achieve high-quality solutions with low inference latency.

\subsection{Diffusion Models}

Discrete diffusion models \cite{song2020} operate in the discrete domain using noise distributions like Bernoulli in the forward process. These models transform the original data distribution into a noised distribution and then train neural networks to reverse this process. They have been effectively applied in graph structure generation \cite{vignac2023}, image generation \cite{song2020}, and improved text generation \cite{chen2023analog}. However, their sampling processes require numerous iterations , which typically results in slow speeds. To address this issue, \citet{song2023} introduced the consistency model, which uses consistency mapping to achieve rapid one-step generation by directly mapping any state in the noise trajectory to its origin.

Recent studies have explored the use of diffusion-based generative models to solve TSP. \citet{graikos2022} extended the diffusion model for image generation to the Euclidean TSP by generating a 64x64 grayscale image to represent the solution of the TSP instance, which is then decoded into a TSP tour. This image-based method is limited as it cannot explicitly model the node/edge selection process, resulting in suboptimal solution quality and increased inference delay. \citet{Sun2023} reformulated the TSP as a discrete \{0,1\} vector optimization problem and utilized a graph-based denoising diffusion model to generate TSP solutions represented by adjacency matrices. \citet{ma2024} proposed a diffusion model training strategy in a semi-supervised manner. \citet{li2024} incorporated objective optimization guidance into the denoising steps through gradient feedback for more efficient sampling. Although these approaches demonstrate improved performance, they rely on the standard DDIM paradigm \cite{song2020}, which necessitates simulating a Markov chain with a substantial number of steps for generation, significantly increasing time overhead. This limitation deprives them of the speed advantage of NAR approaches and hinders their practicality in time-sensitive real-world applications \cite{xu2018}. In contrast, DEITSP directly maps noise to the optimal solution by enhancing self-consistency \cite{song2023} and develops a specialized iteration process tailored for TSP, which contributes to improved efficiency in both inference speed and solution quality. Additionally, DEITSP offers the flexibility to balance solution quality and inference speed by allowing users to choose the number of iteration steps.

\begin{figure}[t]
    \centering
    \includegraphics[width=0.45\textwidth]{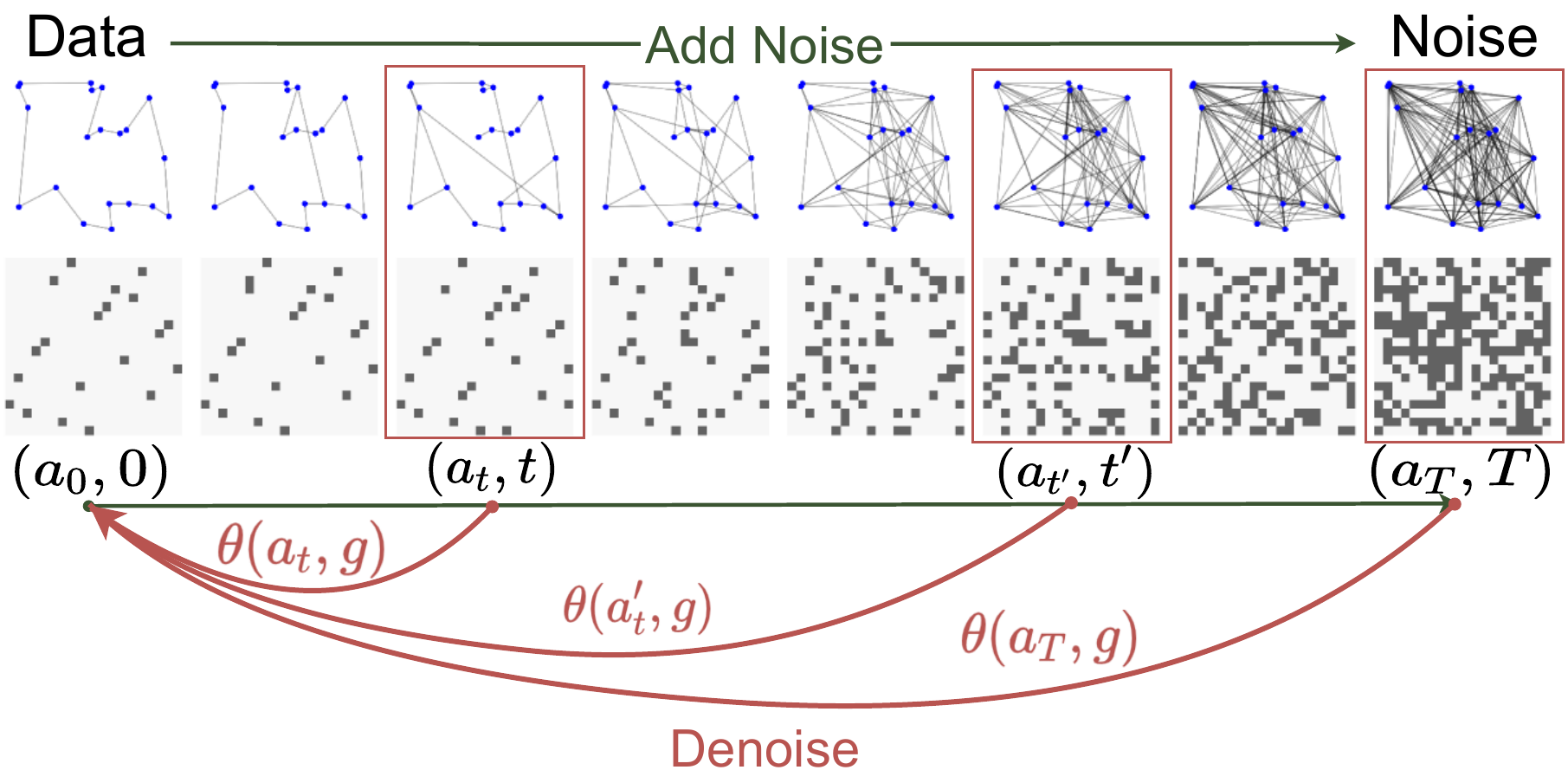}
    \caption{The proposed diffusion model learns to directly map any state in the noise trajectory to its origin state (i.e. the optimal solution), where $\theta$ denotes the neural network and $g$ represents the TSP instance as the condition.}
    \Description{The proposed consistency model learns to directly map any time point in the noise trajectory to its origin time point.}
    \vspace{-5pt}
    \label{fig:self_consistency}
\end{figure}

\section{Methodology}

In this section, we introduce DEITSP, our diffusion-based NAR model designed to efficiently generate high-quality TSP solutions with minimal iteration steps. In Section \ref{diffusion}, we present a one-step diffusion model that combines a controlled discrete noise addition process with self-consistency enhancement, allowing predicting optimal solutions through single-step denoising from various noise data. Section \ref{backbone} details a dual-modality graph transformer, which enhances the extraction and fusion of features from different modalities. Finally, in Section \ref{inference}, we describe an efficient iterative strategy that leverages the network's denoising ability at different noise levels, progressively compressing the solution space, enhancing exploration capabilities, and iteratively improving solution quality, without following the standard DDIM paradigm \cite{song2020}.

\subsection{TSP Setting}
Our research focuses on the two-dimensional Euclidean TSP, widely used and highly significant in various domains. We represent the TSP instance as a graph $g=\left(V, A \right)$ where $V=\{v_i|1\le i\le N\}$ denotes all $N$ node coordinates and $A=\{a_{ij}|1\le i,j\le N\}$ denotes the connectivity between nodes. The connected edges form a Hamiltonian loop that visits each node in $V$ exactly once. The cost of feasible solutions is defined as: $\mathrm{L}(A)= {\textstyle \sum_{i}^{}\textstyle \sum_{j}^{} a_{ij} \times \mathrm{cost} (v_i,v_j)}$, where $\mathrm{cost}(v_i,v_j)$ represents the Euclidean distance between $v_i$ and $v_j$, and $\mathrm{L}(\cdot)$ denotes the length of the TSP solution. For a given TSP instance $s$, our goal is to find the optimal solution that minimizes the cost among feasible ones: 
\begin{equation}
  A^* = \mathrm{arg}\min_{A\in\mathcal{ A}}  \mathrm{L}(A).  
\end{equation}

\subsection{One-step Diffusion Model}
\label{diffusion}
In this subsection, we first introduce the parameterization of the controlled discrete noise addition process. Next, we describe the self-consistency enhancement during the training process, enabling direct prediction of high-quality solutions through single-step denoising from various noise levels.

\subsubsection{Controlled Discrete Noise Addition}

We design a diffusion model based on the discrete diffusion model \cite{austin2021} and consistency model \cite{song2023} to support a controllable process for disturbing TSP solutions with varying noise levels. Similar to image-based diffusion models treating each pixel as data, we consider each edge's connectivity as data following the Bernoulli distribution. Given the optimal solution $A_0$, each edge is represented by a one-hot row vector $a_0\in{0,1}$.

The noise addition process is determined by the transition matrix $\left [ \mathbf{Q_t}\right ] _{ss^{\prime}} = {q}(a_t=s^{\prime}|a_{t-1}=s)$, where $\left [ \mathbf{Q_t}\right ] _{ss^{\prime}}$ represents the probability of transition from state $s$ to state $s^{\prime}$ and $t$ represents the noise level. For any edge, the noise addition process is defined as:
\begin{equation}
    \label{addnoise}
    {q}(a_t|a_{0})=\mathrm{Cate} (a_t;\boldsymbol{p}=a_{0}\mathbf{\overline{Q}_t}),
\end{equation}
where $\mathbf{\overline{Q}_{t}}=\mathbf{Q_1}\mathbf{Q_2}...\mathbf{Q_t}$ and $\mathrm{Cate}(a; p)$ is a categorical distribution over the one-hot row vector $a$ with probabilities $p$ to add noise to disturb an edge. 
We parameterize the probability transition matrix as 
$\mathbf{Q_t}=\begin{bmatrix}
 (1-\beta_t) & \beta_t\\
  \beta_t & (1-\beta-t)
\end{bmatrix}$ following \cite{austin2021}.
The diffusion rates $\beta_t$ should be defined such that $\prod_{t=1}^{T} (1-\beta_t) \approx 0$, ensuring ${q}(a_T|a_0)$ to converge to a uniform distribution independent of $a_0$, i.e., ${q}(a_t|a_0) \approx {Uniform}(a_t) $.  We set $\beta_t$ to follow a linear schedule starting from $\beta_1 = 10^{-4}$ and ending at $\beta_T = 0.02$ as described in \cite{ho2020, graikos2022}.

\begin{algorithm}[tb]
    \caption{Training DEITSP}
    \label{alg:algorithm1}
    \begin{algorithmic}[1]
        \STATE \textbf{Input}: dataset D, initial model parameter $\theta$. 
        \REPEAT
        \STATE $a_0,g \sim D$. \COMMENT{Get TSP instance $g$ and ground truth $a_0$}
        \STATE $t \sim \operatorname{Uniform}(\{1, \ldots, T\})$. 
        \STATE $a_t \sim Cate(a_t;p=a_0\bar{Q_t})$. \COMMENT{Denoise}
        \STATE $a_{t+k} \sim Cate(a_{t+k};p=a_0\bar{Q}_{t+k})$. \\ \COMMENT{Denoise from another noise level}
        \STATE $\textrm{Minimize }\mathcal{L}(a_0,\theta (a_{t},g),\theta (a_{t+k},g)$. 
        \UNTIL{convergence}
    \end{algorithmic}
\end{algorithm}

\subsubsection{Self-consistency Enhancement}

The neural network $\theta$ is trained to directly predict the probability distribution of the optimal solution, namely $\tilde{a}_0$, also known as heatmap scores \cite{Joshi2019, Sun2023}. To facilitate the neural network's learning for one-step denoising, we employ self-consistency \cite{song2023} to enhance the training process. As proposed by consistency models \cite{song2023}, self-consistency ensures that all states along a noise trajectory map back to the same original state, as illustrated in Figure \ref{fig:self_consistency}. This method imposes two key constraints: first, at any state on the noise trajectory, the denoising output generated by the neural network $\theta$ must be consistent with the ground truth, i.e., $\theta (a_t,g)= a_0$ for all $t\in[1,{T}]$, where $g$ represents the input TSP instance as the condition. Second, the denoising outputs from any two states along the noise trajectory must be consistent with each other, i.e., $\theta (a_{t_1},g)  =\theta (a_{t_2},g)$ for all $t_1$, $t_2\in[1,{T}]$.

\subsubsection{Training Process}

We optimize the model through supervised learning, with the ground truth $a_0$ obtained by the exact solver Concorde \cite{Applegate2007}. We add various levels of noise to the ground truth to obtain noisy data $a_t, a_t'$, and input them into the neural network to predict the denoised data $\tilde{a}_0$, $\tilde{a}_0'$'. Then we train the model parameters end-to-end by minimizing the loss via gradient descent. The loss function, defined as Eq. (\ref{eq:loss}), comprises three terms: the first two terms ensure the network's output is consistent with the ground truth, while the third term enforces self-consistency between any two states on the noise trajectory. A parameter $\lambda$ balances these objectives. The overall training process of DEITSP is outlined in Algorithm \ref{alg:algorithm1}. To stabilize training, the loss function ensures consistency not between any time steps, but between two intervals of $k$ steps, i.e., $a_{t+k}$ and $a_t$. Following \cite{luo2023}, we set $k=20$.
\vspace{-2pt}
\begin{align}
    \mathcal{L} = ~& \mathrm{CrossEntropy}(a_0,\theta (a_{t+k},g)) \nonumber 
     + \mathrm{CrossEntropy}(a_0,\theta (a_{t},g)) \\
    & + \lambda \cdot ||\theta (a_{t+k},g)-\theta (a_{t},g)||_2 .  \label{eq:loss}
\end{align}

\subsection{Dual-modality Graph Transformer}
\label{backbone}

In this section, we describe the three key components of our proposed dual-modality graph transformer. Existing methods \cite{Joshi2019, Fu2021, Sun2023} typically utilize the anisotropic graph neural network with edge-gating mechanisms to extract latent features of the TSP through local aggregation. But their limited receptive field hinders the full exploitation of available data and restricts network depth, leading to slow inference speeds, particularly during multi-step denoising iterations. Moreover, deep GNNs suffer from excessive smoothing due to repeated local aggregation \cite{hussain2022}. 

\begin{figure}[t]
    \centering
    \includegraphics[width=0.40\textwidth]{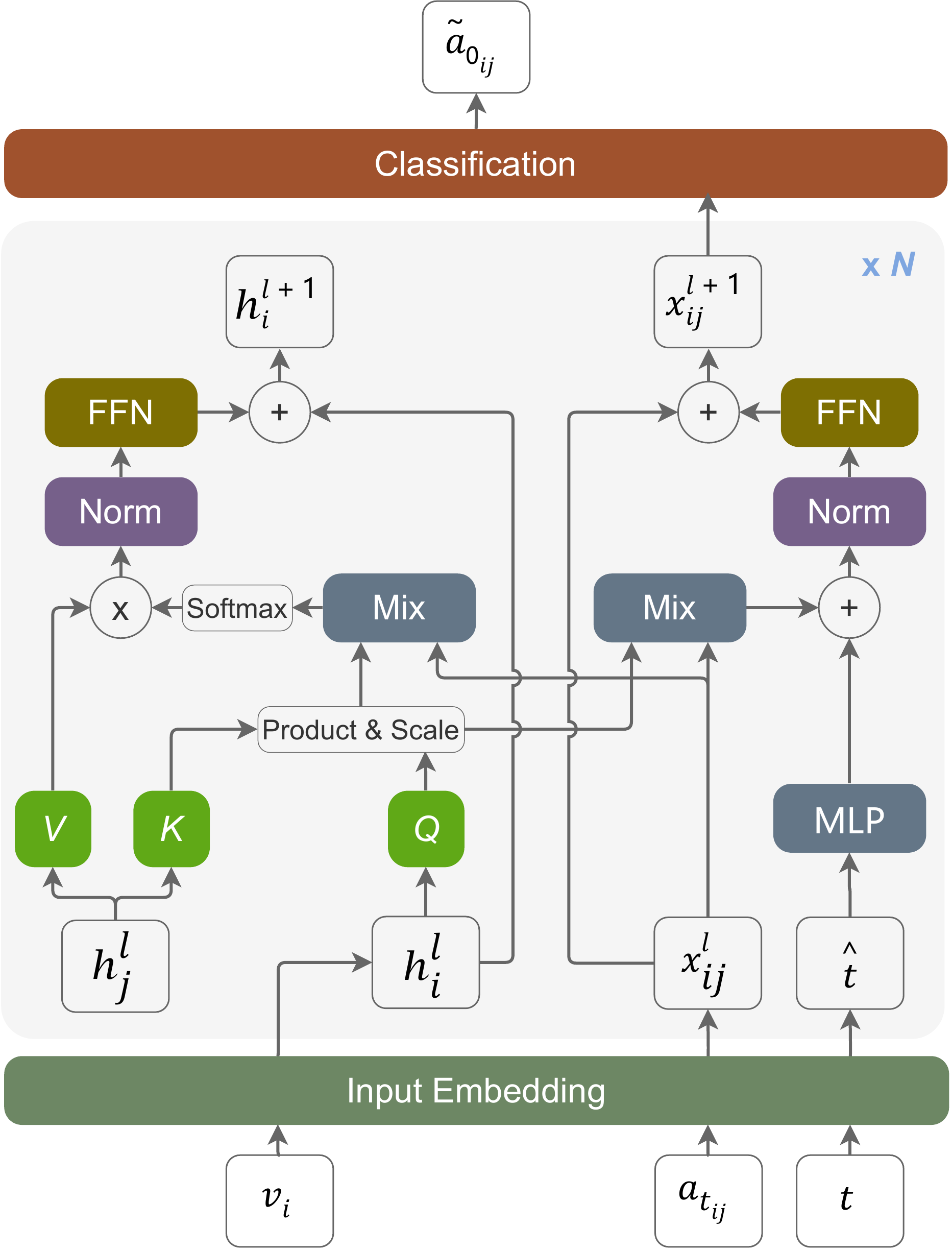}
    \caption{The network takes node coordinates $v_i$, noisy adjacency matrix $a_{t_{ij}} $ and time step $t$ as inputs to predicts the heatmaps $\tilde{a}_{0_{ij}}$. Mix refers to Eq. (\ref{eq:mix1}) and (\ref{eq:mix2}), while FFN and Norm is Feed Forward and Layer Normalization used in \protect\citet{Vaswani2017}}
    \Description{The parametrization of our proposed graph transformer.}
    \vspace{-8pt} 
    \label{fig:network}
\end{figure}

To address these limitations, we design a dual-modality network based on the attention mechanism for continuous modes of nodes and discrete modes of edge links. Unlike neighborhood aggregation in conventional graph neural networks, the attention mechanism allows multi-hop interactions, thereby capturing a broader scope of structural information and reducing dependence on neural network depth. The personalized information of nodes is retained by applying different attention weights, which reduces the risk of node representations becoming overly similar. \citet{wu2024demystifying} also indicates that the attention mechanism demonstrate slower rates of oversmoothing compared to GCNs. The backbone of DEITSP (Figure \ref{fig:network}) consists of three key components: 1) a linear layer for input embedding; 2) multiple dual-modal learning layers for facilitating information transfer among nodes and edges; and 3) several classification layers for outputting edge scores.

\subsubsection{Input Layer}

For the given TSP instance $g=(V,A_t)$, $A_t=\{a_{t_{ij}}|1\le i,j\le N\}$ represents the noisy adjacency matrix and $V=\{v_i|1\le i\le N\}$ represents the node coordinates. DEITSP linearly projects each input node $v_i$, edge $a_{t_{ij}}$, and time step $t$ that indicates noise level into $d$-dimensional hidden features. Their initial embedding is computed as follows:
\begin{align}
    h_i^0 = & \mathbf{W_v} v_i+\mathbf{b_v}+\mathrm{PE},\nonumber \\
    x_{ij}^0 = & \mathbf{W_a} a_{t_{ij}}+\mathbf{b_a}+\mathrm{PE},  \\
    \hat{t} = & \mathbf{W_t} t+\mathbf{b_t}+\mathrm{PE}, \nonumber
\end{align}
where $\mathbf{W_v}$, $\mathbf{W_a}$ and $\mathbf{b_v},\mathbf{b_a},\mathbf{b_t}$ are the parameters of the linear projection layers, and PE is the positional encoding \cite{Vaswani2017}.

\subsubsection{Dual-modal Learning (DML) Layer}

DEITSP utilizes $N$ dual-modal learning layer to extract information from TSP instances and maintain the features of nodes and edges at each layer. Edge features represent discrete adjacency relationships, while node features represent continuous node aggregation. Considering different modalities, we conducted separate feature extraction and mixing. 

For nodes, we first use a mixing module to integrate edge features $x_{ij}^{l}$ into the implicit edge scores $\hat{w}_{ij}^{l}$ calculated from paired attention between nodes as in Eq. (\ref{eq:mix1}). Secondly, we use the scores $w_{ij}^{l}$ combining edge features to aggregate node features as in Eq. (\ref{eq:node}), and transmit them to the normalization layer, feedforward network, and residual connection to update node features as in Eq. (\ref{eq:ffn1}).  

\begin{align}
     w_{ij}^{l} = & \mathbf{W_{e_1}} x_{ij}^l \odot{}\hat{w}_{ij}^l + \mathbf{W_{e_2}} x_{ij}^l+ \hat{w}_{ij}^l\label{eq:mix1}, \\
    \hat{h}_i^{l+1} = & \mathrm{softmax} ({w_{ij}^l}) \times{} \mathbf{V^l}h_j\label{eq:node}, \\
    h _i^{l+1} = & \mathbf{W_{h_2}}(\text{ReLU}( \mathbf{W_{h_1}}(\mathrm{Norm} (\hat{h} _i^{l+1}))))+h_{i}^l\label{eq:ffn1},
\end{align}
where $l$ is the layer index and $V^l$ represents the weight for calculating attention at each layer. $\hat{w}_{ij}^l=(\frac{\mathbf{Q^l}h_i^l  (\mathbf{K^l} h_j^l)^T }{\sqrt{d} } )$ represents the implicit edge scores calculated from paired attention between nodes.

For edges, we first use a mixing module to fuse the implicit edge scores $\hat{w}_{ij}^{l}$ into edge features $x_{ij}^{l}$ as in Eq. (\ref{eq:mix2}). The mixing modules of node and edge operate independently, distinguishing from typical graph transformers. Secondly, we combine time step features $\hat{t}$ with the mixed edge features $\hat{x}_{ij}^{l}$ as in Eq. (\ref{eq:edge}), and pass them through the normalization layer, feedforward network, and residual connection to update edge features in Eq. (\ref{eq:ffn2}) as:

\begin{align}
    \hat{x}_{ij}^{l+1} = & \mathbf{W_{y_1}} \hat{w}_{ij}^l \odot{}x_{ij}^l + \mathbf{W_{y_2}} \hat{w}_{ij}^l+ x_{ij}^l\label{eq:mix2}, \\
    \hat{x} _{ij}^{l+1}= & \hat{x} _{ij}^{l+1} + \text{MLP}(\hat{t})\label{eq:edge}, \\
    x _{ij}^{l+1} = & \mathbf{W_{x_2}}(\text{ReLU}( \mathbf{W_{x_1}}(\mathrm{Norm} (\hat{x} _{ij}^{l+1}))))+x_{ij}^l\label{eq:ffn2}.
\end{align}
where $l$ is the layer index and $\text{MLP}(\hat{t})=\mathbf{W_{t}^{l}}(\alpha (\hat{t}))$. 

\subsubsection{Classifier Layer}

The classifier layer takes the edge embedding $x_{ij}^{N+1}$ of the last layer as input to predicts the heatmaps $\tilde{a}_{0_{ij}}$ of the optimal solution as follows:

\begin{equation}\label{eq:classfier}
    \tilde{a}_{0_{ij}}=\mathrm{Conv1d}(\alpha (\mathrm{Conv1d}(\mathrm{GN}(x_{ij}^{N+1})))),
\end{equation}
where $\mathrm{Conv1d}$ is 1-d convolution, $\mathrm{GN}$ is the group normalization and $\alpha$ is the sigmoid activation function.

\subsection{Iterative Denoising and Noise Scheduling}
\label{inference}

In this section, we present the inference process of DEITSP. First, we use a greedy strategy to decode the heatmaps predicted by neural networks into feasible TSP tours. A specific introduction to the decoding strategy is provided in Appendix \ref{decoding}. Second, we propose an efficient iterative strategy that leverages denoising at different noise levels to enhance exploration. Finally, we design a novel schedule to control the level of noise addition during iterations, thereby gradually refining the solution space.

\subsubsection{Iterative Strategy to Enhance Exploration}

Unlike image generation, which has no fixed optimal solution, TSP has well-defined optimal solutions, making deviations during denoising less tolerable. The large solution space also complicates finding the optimal solution. Existing diffusion-based methods that employ the DDIM paradigm \cite{song2020} aim for a single high-quality solution through many generation steps. However, given the specific nature of TSP, which requires precise solutions and efficient exploration of a large solution space, previous approaches may not be effective for TSP.

\begin{algorithm}[t]
    \caption{DEITSP Inference}
    \label{alg:algorithm2}
    \begin{algorithmic}[1]
        \STATE \textbf{Input}: Diffusion model $\theta$, noise schedule $\tau_i$, iteration step $M$, TSP instance $g$.
        \STATE $a_t \sim \operatorname{Uniform}(a_t),\;t = T$ \COMMENT{Generate initial solution}
        \FOR{$1$ \textbf{ to }  $M$ }
    		\STATE $\tilde{a_0} = \theta (a_{t},g)$. \COMMENT{One step denoising}
    		\STATE $\tilde{A_0} \leftarrow  \tilde{a_0} $ \COMMENT{Aggregate to solutions set}
    		\STATE $t=\tau_i$
    		\STATE $a_t \sim \mathrm{Cate} (a_t;p=a_0 \mathbf{\overline{Q}_{t}})$ \COMMENT{Adding noise}
	\ENDFOR
	\STATE ${a_0}^* = \arg \min_{\tilde{a_0}} (\text{GreedyDecoding} (\tilde{A_0}))$ \\ \COMMENT{Get the best solution from solutions set}
	\STATE \textbf{Output}: ${a_0}^*$ 
    \end{algorithmic}
\end{algorithm}

Different from the image generation task, which is challenging to achieve through single-step prediction, existing NAR models \cite{Joshi2019, Xiao2023} have demonstrated the ability to effectively learn the latent representation of TSP solutions through single-step prediction. Consequently, we design a one-step diffusion model that directly maps noise to high-quality solutions through single-step denoising, bypassing the need for multi-step Markov processes.

\begin{figure}[t]
    \centering
    \includegraphics[width=0.45\textwidth]{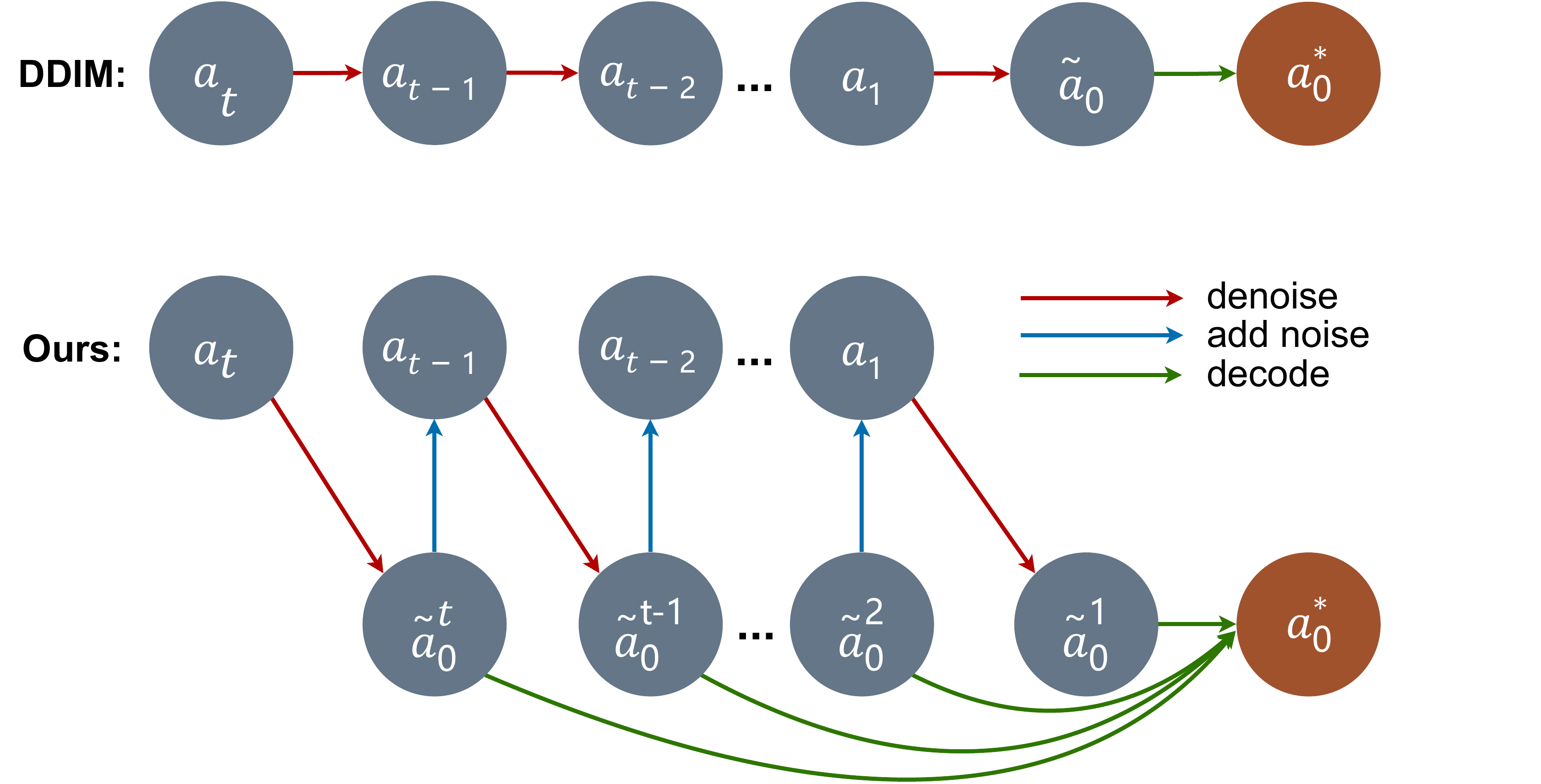}
    \caption{DDIM iteratively predicts $a_{t-1}$ to progress the chain of mapping noise into data, discarding previously estimated $\tilde{a_0}$. In contrast, our method aggregates the outputs of the diffusion model at multiple steps and decodes all outputs $\tilde{a}_0^t$ to improve solution quality.}
    \Description{Our method aggregates the outputs of multiple steps and decodes all outputs instead of the last one to explore higher-quality solutions.}
    \label{fig:inference}
\end{figure}

Additionally, to reduce errors, DEITSP does not limit itself to a single high-quality solution but instead explores multiple solutions. We design an iterative strategy that alternates between adding and removing noise to enhance exploration. As shown in Figure \ref{fig:inference}, we add noise to the denoised data from the previous time step and denoise it again, leveraging different noise levels to obtain new results. We then aggregate the results $\{\tilde{a}_0^t \mid t \in [1, T]\}$ and decode all of them using greedy decoding to generate multiple solutions. This differs from the typical DDIM approach, where previously estimated results like $a_{t}$ are discarded when estimating $a_{t-1}$ at a new time step. Our iteration process fully utilizes the outputs from all iteration steps, exploring more high-quality solutions with minimal extra decoding cost. Moreover, unlike multiple sampling, where the inputs for each iteration are randomly generated, our method confines the exploration to relevant local regions of denoising data $\{\tilde{a}_0^t \mid t \in [1, T]\}$, improving the efficiency of the exploration. Notice that DEITSP is flexible to generate solutions from a single time step ($T=1$) or multiple time steps ($T>1$).

\begin{table*}[!t]
  \centering
  \caption{Performance of DEITSP with comparisons with benchmarking methods on TSP20, TSP50 and TSP100.}
  \label{tab:comparison}
  \resizebox{1.8\columnwidth}{!}{
  \begin{tabular}{lcccccccccc}
  \toprule
  \multirow{2}{*}{\textbf{Methods}}                & \multirow{2}{*}{\textbf{Type}} & \multicolumn{3}{c}{TSP20}                    & \multicolumn{3}{c}{TSP50}                    & \multicolumn{3}{c}{TSP100}                   \\
                                         &                       & Len  & Gap(\%)  & Time    & Len  & Gap(\%)  & Time    & Len  & Gap(\%)  & Time    \\
  \midrule
  Concorde          & Exact           & 3.84           & 0.00               & -               & 5.69           & 0.00               & -               & 7.76           & 0.00               & -      \\
  2 opt             & Heuristic       & 3.94           & 2.51               & 8.41s           & 6.02           & 5.89               & 24.54s          & 8.35           & 7.61               & 53.74s \\
  \midrule
  GAT*              & NI, S, 2OPT     & 3.84           & 0.09               & 6m              & 5.75           & 1.00               & 32m             & 8.12           & 4.64               & 5h     \\
  DACT(1k Iter)     & NI, G           & 3.84           & 0.01               & 17.86s          & 5.69           & 0.11               & 43.38s          & 7.89                & 1.62               & 1.75m  \\
  \midrule
  AM               & AR, BS1280       & \underline{3.84} & \underline{0.00} & 32.17s          & 5.70          & 0.25          & 1.44m           & 7.95          & 2.44          & 3.61m           \\
  TRANSFORMER\dag  & AR, BS1000       & 3.85          & 0.09          & 46.24s          & \underline{5.69}     & \underline{0.02}         & 2.89m           & 7.80                & 0.46               & 7.49m          \\
  MDAM             & AR, G            & 3.86          & 0.57          & 4.69m           & 5.78          & 1.56          & 13.68m          & 7.99          & 2.99          & 20.38m          \\
  LCP(20 Iter)     & AR, S1280        & 3.84          & 0.00          & 16.69m          & 5.69          & 0.02          & 23.95m          & 7.80          & 0.55          & 43.43m          \\
  POMO             & AR, G, AUG       & 3.84          & 0.00          & 21.71s          & 5.69          & 0.09          & 44.84s          & 7.77          & 0.19          & 1.26m           \\
  Sym-NCO          & AR, G, AUG       & 3.85          & 0.08          & 22.44s          & 5.69          & 0.04          & 45.97s          & \underline{7.77} & \underline{0.18} & 1.79m     \\
  CNN\_Transformer\dag & AR, BS1000   & 3.85          & 0.28          & 50.01s           & 5.69          & 0.10          & 3.39m           & 7.85          & 1.13          & 9.99m           \\
  Pointerformer    & AR, G, AUG      & 3.84          & 0.01          & 38.58s          & 5.69          & 0.10          & 1.35m           & 7.77          & 0.12          & 2.29m           \\
  ELG\dag          & AR, G, AUG       & 3.85          & 0.10          & 55.22s          & 5.69          & 0.07          & 2.18m           & 7.77          & 0.20          & 4.14m           \\
  HierTSP          & AR, G, AUG       & 3.84          & 0.00          & 33.48s          & 5.69          & 0.03          & 1.92m           & 7.78          & 0.21          & 2.79m           \\
  \midrule
  GCN              & NAR, BS1280      & 3.87          & 0.80          & 27.40s          & 5.72          & 0.53          & 55.62s          & 7.96          & 2.53          & 1.78m           \\
  NAR4TSP\dag      & NAR, BS1000      & 3.85          & 0.29          & \underline{16.36s}  & 5.70      & 0.25          & \underline{27.21s}  & 7.82      & 0.81          & \underline{44.15s} \\
  Image Diffusion* & NAR, G, 2OPT     & -             & -             & -               & 5.76          & 1.23          & -               & 7.92          & 2.11          & -               \\
  DIFUSCO\dag      & NAR, G, 2OPT     & 3.85          & 0.19          & 6.88m           & 5.69          & 0.09          & 7.56m           & 7.78          & 0.23          & 9.47m           \\
  T2TCO\dag        & NAR, G,2OPT      & 3.84          & 0.05          & 11.18m          & 5.69          & 0.02          & 24.91m           & 7.77          & 0.13          & 26.46m           \\
  DEITSP(1 Iter)   & NAR, G, 2OPT  & 3.84          & 0.01          & \textbf{10.40s} & 5.70          & 0.12          & \textbf{15.05s} & 7.81          & 0.63          & \textbf{30.82s} \\
  DEITSP(4 Iter)   & NAR, G, 2OPT  & 3.84          & 0.00          & 27.48s          & 5.69          & 0.03          & 32.37s          & 7.78          & 0.21          & 1.13m           \\
  DEITSP(16 Iter)  & NAR, G, 2OPT & \textbf{3.84} & \textbf{0.00} & 1.63m           & \textbf{5.69} & \textbf{0.01} & 1.99m           & \textbf{7.77} & \textbf{0.10} & 3.99m          \\ 
  \bottomrule
  \multicolumn{11}{p{1.82\columnwidth}}{G, S, BS, 2OPT, and AUG represent Greedy-search, Sampling, Beam-search, 2-opt process, and instance augmentation, respectively. * denotes baseline results from the original paper. \dag~ indicates testing on TSP20/50 using pre-trained TSP50 weights provided by the original paper. \textbf{Bold} indicates the best solution; \underline{underlined} indicates the second-best solution.}
  \end{tabular}}
\end{table*}

\subsubsection{Noise Schedule to Refine the Solution Space}

The level of noise added in each iteration affects both the exploration range and the solution space of the denoising process. Higher noise levels increase the exploration range but also introduce errors in network prediction. To balance this, we adopt a decreasing noise addition schedule that enhances exploration while gradually refining the solution space. Formally, we define the level of noise $\tau$ added in each iteration as a descending sequence within the range from $T$ to $1$. The setting of $\tau$ can be flexible, such as the commonly used linear schedule $\tau_i = \lfloor c_i \times T \rfloor$ for $c_i$ with uniform values in [0,1], or the cosine schedule $\tau_i = \lfloor \mathrm{cos}(\frac{(1-c_i)\pi}{2}) \times T \rfloor$ for $c_i$ with uniform values in [0,1]. A more general definition can also be employed:
\begin{equation}
    \tau_i=\left \lfloor (\frac{f(c_i)-\mathrm{min}(f(c_i))}{\mathrm{max}(f(c_i))-\mathrm{min}(f(c_i))} )\times T \right \rfloor ,
\end{equation}
where $f(\cdot)$ represents any elementary function, and $c_i$ can be uniformly distributed over any interval $[l,r]$. In this paper, we use the inverse function as $f(\cdot)$, with $c_i$ uniformly distributed in the interval $[0.25, 1.5]$. This choice is motivated by the sparsity of the optimal solution, which leads to a significant difference in the distribution between the high-noise state and the optimal solution. Therefore, unlike the linear and cosine schedules typically used in image generations, the TSP requires more iterations in low-noise states. We present experiments on scheduling methods to verify this in Section \ref{ablation} and provide detailed comparisons in Appendix \ref{ap:schedule}.

Algorithm \ref{alg:algorithm2} outlines the overall inference process of DEITSP. The initial solution $a_t$ is sampled from a uniform distribution $Uniform(a_t)$ at time step $t = T$ (Step 2) and then input into the diffusion model $\theta$ to remove noise and predict the heatmaps $\tilde{a}_0$ of the optimal solution (Steps 4 and 5). Multiple solutions are then obtained by alternately adding and removing noise, with the noise schedule $\tau_i$ controlling the level of noise added in each step (Steps 6 and 7). Finally, we employ a greedy decoding method to produce the solutions from all steps and select the highest-quality solution (Steps 9 and 10).\vspace{-1pt}

\section{Experiments}

This section presents comprehensive evaluations of DEITSP. We begin by outlining the experimental settings, including benchmark datasets and baseline models. Then, we compare DEITSP’s performance with baseline models in solution quality and inference efficiency. Next, we assess DEITSP’s generalization capability across TSPs of varying sizes and distributions. Finally, we conduct ablation studies to examine the contributions of DEITSP’s components.

\begin{table*}[t]
  \centering
  \caption{Generalization performance of applying models trained using TSP100 on TSP200, TSP500 and TSP1000.}
  \label{tab:Generalization}
  \resizebox{1.8\columnwidth}{!}{
  \begin{tabular}{lcccccccccc}
  \toprule
  \multirow{2}{*}{\textbf{Methods}}          & \multirow{2}{*}{\textbf{Type}} & \multicolumn{3}{c}{TSP200}                   & \multicolumn{3}{c}{TSP500}                   & \multicolumn{3}{c}{TSP1000}                  \\
                                    &                       & Len & Gap(\%) & Time   & Len & Gap(\%) & Time   & Len & Gap(\%) & Time   \\
  \midrule
  Concorde                                     & Exact                 & 10.72          & 0.00               & 3.44m           & 16.55          & 0.00               & 37.66m          & 23.12          & 0.00               & 6.65h          \\
  GAT*                                         & S, 2OPT               & 11.61          & 8.32               & 9.59m           & 23.75          & 43.57              & 57.76m          & 47.73          & 106.46             & 5.39h          \\
  GCN                                          & BS1280                & 16.19          & 51.02              & 4.63m           & 30.37          & 83.55              & 38.02m          & 51.26          & 121.73             & 51.67m         \\
  AM                                           & BS1280                & 11.38          & 6.14               & 5.77m           & 19.53          & 18.03              & 21.99m          & 29.90          & 29.24              & 1.64h          \\
  TRANSFORMER                                  & BS2500                & 11.14          & 3.96               & 5.66m           & 20.77          & 25.55              & 34.30m          & 33.29          & 44.00              & 2.23h          \\
  POMO                                         & G, AUG              & 10.89          & 1.64               & 20.48s          & 19.52          & 17.98              & \underline{1.33m}  & 30.51          & 31.96              & 8.80m          \\
  Sym-NCO                                      & G, AUG              & 10.90          & 1.64               & \underline{20.14s} & 19.84          & 19.93              & 1.35m           & 31.47          & 36.13              & 8.88m          \\
  Att-GCN*                                     & MCTS                  & 10.81          & 0.88               & 2.50m           & 16.97          & 2.54               & 5.91m           & \textbf{23.86} & \textbf{3.22}      & 12.47m         \\
  DIFUSCO                                      & G, 2OPT               & 10.81          & 0.78               & 2.98m           & 17.13          & 3.51               & 17.28m          & 24.25          & 4.89               & 1.15h          \\
  T2TCO                                        & G,2OPT                & \underline{10.78} & \underline{0.55}      & 6.18m           & 17.01          & 2.78               & 38.40m          & OOM            & OOM                & OOM            \\
  Pointerformer                                & G, AUG              & 10.87          & 1.45               & 26.49s          & 18.62          & 12.53              & 1.35m           & 27.89          & 20.66              & 6.89m \\
  ELG                                          & G, AUG              & 10.90          & 1.64               & 46.64s          & 17.73          & 7.16               & 1.94m           & 25.74          & 11.36              & \underline{3.99m} \\
  HierTSP                                      & G, AUG              & 10.96          & 2.23               & 24.93s          & 18.54          & 12.08              & 1.50m           & 28.39          & 22.79              & 9.43m \\
  BQ-NCO                                       & G                   & 11.20          & 4.51               & 2.52m          & 17.21          & 4.02              & 7.05m           & 24.09          & 4.19              & 23.43m \\
  DEITSP(1 Iter)                               & G, 2OPT           & 10.89          & 1.52               & \textbf{11.40s} & 17.06          & 3.12               & \textbf{53.70s} & 24.14          & 4.41               & \textbf{3.77m} \\
  DEITSP(4 Iter)                               & G, 2OPT           & 10.80          & 0.70               & 24.80s          & \underline{16.96} & \underline{2.51} & 2.55m           & 24.06          & 4.07               & 11.24m         \\
  DEITSP(16 Iter)                      & G, 2OPT          & \textbf{10.77} & \textbf{0.40}      & 1.41m           & \textbf{16.90} & \textbf{2.15}      & 9.31m           & \underline{23.97} & \underline{3.68}      & 41.83m        \\
  \bottomrule
  \multicolumn{11}{l}{\textbf{Bold} indicates the best solution; \underline{underlined} indicates the second-best solution.}
  \end{tabular}}
  \end{table*}

\subsection{Experiment Configurations}
\label{hyperpara}

\hspace{1em}\textbf{Dataset } The training instances for TSP-20/50/100 are identical to those used in previous studies \cite{Kool2019,Kwon2020,Joshi2022}, where instances are uniformly generated within the range $[0,1]$. We generate the same test instances as \cite{Joshi2022} for TSP-20/50/100 and as \cite{Fu2021} for TSP-200/500/1000. To verify the generalization performance on realistic data, we sample instances from city maps of three different real-world countries, namely USA13509, BM33708 and JA9847 \cite{Yong2024}, as well as TSPLIB \cite{reinelt1991tsplib}. We randomly sample problems of size 100 from the countries map. The optimal solution of testing instances are attained via Concorde \cite{Applegate2007}.

\textbf{Baselines }  We conduct extensive comparisons of DEITSP against various baselines, including the exact solver Concorde \cite{Applegate2007}, heuristic method 2-opt \cite{2opt}, and several state-of-the-art learning-based approaches. The learning-based methods evaluated are categorized into three recent advancements: neural improvement methods, AR methods, and NAR methods. For neural improvement methods, we compare DEITSP with GAT \cite{Deudon2018} and DACT \cite{Ma2021}. For AR methods, the baselines include AM \cite{Kool2019}, Transformer \cite{Bresson2021}, MDAM \cite{Kim2021}, LCP \cite{Xin2021}, POMO \cite{Kwon2020}, Sym NCO \cite{kim2022}, CNN\_Transformer \cite{Jung2023}, Pointerformer \cite{jin2023}, ELG \cite{Gao2023}, BQ-NCO \cite{drakulic2024bq}, and HierTSP \cite{Yong2024}. For NAR methods, we compare DEITSP with GCN \cite{Joshi2019}, Att-GCN \cite{Fu2021}, NAR4TSP \cite{Xiao2023}, and diffusion-based approaches such as Image Diffusion \cite{graikos2022}, DIFUSCO \cite{Sun2023}, and T2TCO \cite{li2024}. We use three key metrics for evaluation: average tour length (Len), average relative performance gap to Concorde (Gap), and total run time (Time).

\textbf{Model Setting } 
We use the same hyperparameters for DEITSP across all experiments. The graph transformer includes 6 dual-modal learning layers, each with a hidden dimension of 256 and 8 attention heads. For noise addition, we set $T=1000$ with a linear noise schedule for ${\beta}_t$ where $\beta_1={10}^{-4}$ and $\beta_T=0.02$. During inference, we use an inverse function iteration schedule, focusing on 1, 4, and 16 iteration steps. To ensure fair comparison, all NN-based models use the same batch size, with other settings following their defaults.
All experiments are conducted on a system with an Intel Xeon Gold 6254 CPU and an NVIDIA RTX 3090 GPU. Our code and datasets will be publicly available.

\begin{table*}[!t]
  \centering
  \caption{Generalization performance of applying models trained using TSP100 instances onto real-world instances.}
  \label{tab:real}
  \resizebox{\textwidth}{!}{
  \begin{tabular}{lcccccccccccc}
  \toprule
  \multirow{2}{*}{\textbf{Methods}} & \multicolumn{3}{c}{JA9857} & \multicolumn{3}{c}{BM33708} & \multicolumn{3}{c}{USA13509} & \multicolumn{3}{c}{TSPLIB} \\
    & Len & Gap(\%) & Time & Len & Gap(\%) & Time & Len & Gap(\%) & Time & Len & Gap(\%) & Time \\
  \midrule
  Concorde       & 78504.28 & 0.00 & -    & 64760.71 & 0.00 & -    & 2113007.65 & 0.00 & -    & 31466.12 & 0.00 & -     \\
  AM             & 506106.19 & 544.69 & 3.50m & 283475.19 & 337.73 & 3.51m & 10344162.00 & 389.55 & 3.58m & 32704.00 & 3.93 & 13.15s \\
  POMO           & 86095.99 & 9.67 & 1.27m & 67152.81 & 3.69 & 1.31m & 2252997.26 & 6.63 & 1.29m & 31902.33 & 1.39 & \underline{2.91s} \\
  Sym-NCO        & 86011.14 & 9.56 & 1.79m & 66326.07 & 2.42 & 1.89m & 2234747.53 & 5.76 & 1.85m & 32069.48 & 1.92 & 2.98s \\
  DIFUSCO        & 80360.48 & 2.36 & 9.53m & 67862.36 & 4.79 & 9.47m & 2202552.75 & 4.24 & 9.49m & 31870.25 & 1.28 & 20.56s \\
  T2TCO          & \underline{79465.05} & \underline{1.22} & 26.48m & 66259.21 & 2.31 & 26.52m & 2159342.20 & 2.19 & 26.46m & \underline{31750.49} & \underline{0.90} & 28.96s \\
  ELG            & 82528.93 & 5.13 & 4.14m & \textbf{65765.26} & \textbf{1.55} & 4.21m & 2158436.50 & 2.15 & 4.17m & 31825.52 & 1.14 & 6.74s \\
  HierTSP        & 84640.31 & 7.82 & 2.79m & 66375.73 & 2.49 & 2.97m & \underline{2157443.98} & \underline{2.10} & 3.58m & 32025.60 & 1.78 & 3.86s \\
  DEITSP(1 Iter) & 80222.46 & 2.19 & \textbf{31.83s} & 67793.80 & 4.68 & \textbf{32.25s} & 2199481.21 & 4.09 & \textbf{35.42s} & 32000.50 & 1.70 & \textbf{2.61s} \\
  DEITSP(4 Iter) & 79656.97 & 1.47 & \underline{1.18m} & 67010.29 & 3.47 & \underline{1.15m} & 2175058.13 & 2.94 & \underline{1.22m} & 31777.83 & 0.99 & 4.82s \\
  DEITSP(16 Iter) & \textbf{79173.08} & \textbf{0.85} & 4.03m & \underline{66249.02} & \underline{2.30} & 3.99m & \textbf{2152979.27} & \textbf{1.89} & 4.11m & \textbf{31711.94} & \textbf{0.78} & 8.51s \\
  \bottomrule
  \multicolumn{13}{l}{\textbf{Bold} indicates the best solution; \underline{underlined} indicates the second-best solution.}
  \end{tabular}}
\end{table*}

\subsection{Performance Comparison}
\label{comparison}

Table \ref{tab:comparison} highlights the superior performance of DEITSP in both solution quality and inference speed. Specifically, with 16 iteration steps, DEITSP achieves results with minimal gaps to the optimal solutions produced by the exact algorithm Concorde. Compared to diffusion-based models like T2TCO \cite{li2024} and DIFUSCO \cite{Sun2023}, DEITSP demonstrates a significant improvement in inference speed, achieving approximately 13 times faster inference on TSP50 instances and 7 times faster on TSP100 instances. Remarkably, with just one iteration step, DEITSP maintains an absolute advantage in inference speed over all learning-based models while delivering competitive solution quality. When compared to the TRANSFORMER model \cite{Bresson2021}, which delivers the best solution quality on TSP50 instances, DEITSP offers substantial speed advantages, with approximately 18 times faster inference on TSP50 and 28 times faster on TSP100, while only experiencing a slight reduction in solution quality (approximately $0.18\%$ on TSP50 and $0.26\%$ on TSP100). Overall, DEITSP outperforms all other NN-based methods in solution quality with minimal inference time, underscoring its ability to produce high-quality solutions with low latency. Furthermore, DEITSP provides the flexibility to balance solution quality and inference speed by allowing the adjustment of iteration steps.

\subsection{Generalization to Larger Instances}

Table \ref{tab:Generalization} presents the performance of models trained on TSP100 instances and tested on 128 TSP200/500/1000 instances. Notably, DEITSP, with just one iteration step, achieves the fastest inference time while maintaining excellent solution quality. Increasing the iteration steps to 16 allows DEITSP to deliver the best performance on TSP200 and TSP500 instances, and a close second on TSP1000 instances, trailing the top-performing Att-GCN \cite{Fu2021} by a small margin (approximately 0.46\%).  While Att-GCN is tailored for large instances, its reliance on Monte Carlo tree search makes it inherently time-consuming. In contrast, DEITSP offers a more flexible approach, allowing users to adjust the number of iteration steps to balance between computation time and solution quality. The results in Table \ref{tab:Generalization} also indicate that DEITSP's generalization performance improves as the number of iteration steps increases, making it a robust and versatile choice across different TSP instance sizes.

\subsection{Generalization to Real-world Instances}

In Table \ref{tab:real}, the node coordinates in the test instances are all sampled from a uniform distribution, consistent with the training data, which may not accurately represent real-world applications. For example, real-world datasets such as TSPLib \cite{reinelt1991tsplib} contain instances with distinct geographical properties that do not satisfy uniform distribution. To verify performance on unseen distributions, we sample locations from real-world maps of cities in the USA, Japan, and Burma \cite{reinelt1991tsplib, Yong2024}. Table \ref{tab:real} compares the performance of different models on TSP100 instances drawn from these countries. Our model demonstrates a clear advantage in the USA and Japan, with a narrower margin in Burma. Additionally, we observe that the performance of learning-based models tends to be unstable on real-world instances of the same size but different distributions. This suggests that previous learning-based models may have overlooked the importance of generalization to real-world scenarios. In contrast, our approach generates multiple solutions from different noise levels, significantly improving robustness. We further evaluate generalization using the widely adopted TSPLIB \cite{reinelt1991tsplib} dataset, which serves as a benchmark for testing solvers on unseen problems. For this evaluation, we randomly select 26 real-world TSP instances of varying sizes (ranging from 51 to 200 nodes) from the TSPLIB dataset as test cases. As shown in Table \ref{tab:real}, DEITSP outperforms all other models on the TSPLIB dataset, further validating the practicality of our method. Full results for each case are provided in Appendix \ref{app:tsplib}.

\begin{figure}[t]
    \centering
    \includegraphics[width=0.45\textwidth]{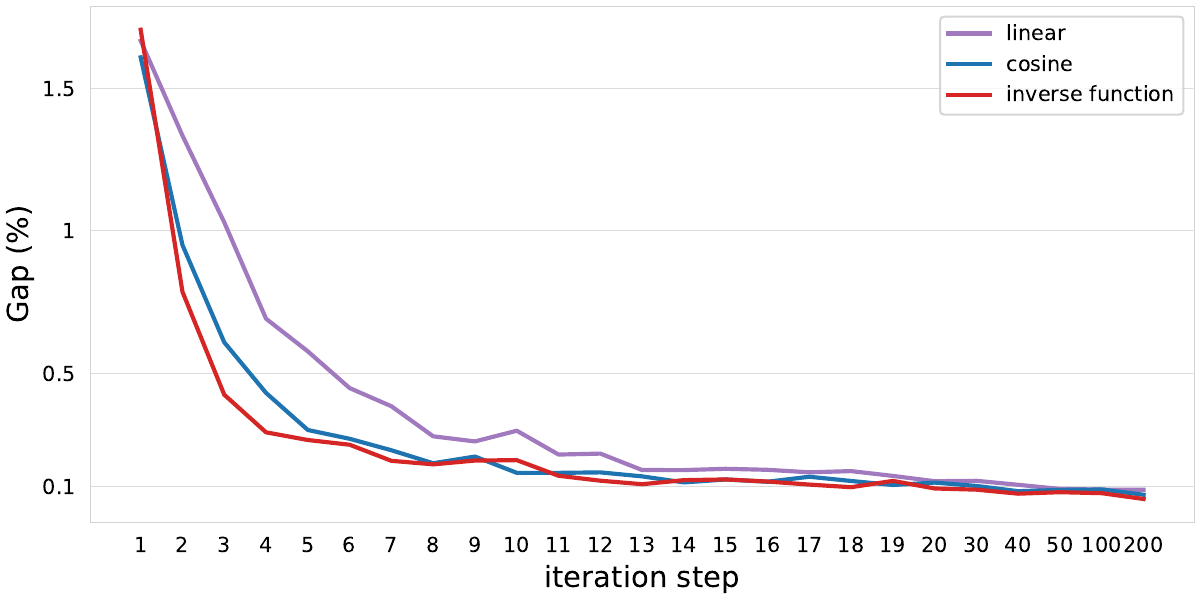}
    \caption{Comparison of the performance of DEITSP with three types of noise schedules and various iteration steps.}
    \Description{Comparison of the performance of DEITSP with three types of noise schedules and various iteration steps.}
    \label{fig:ablation_step}
\end{figure}
\subsection{Ablation Studies}
\label{ablation}

In this subsection, we conduct a series of ablation experiments to evaluate the effectiveness of our proposed graph transformer and iteration strategy on 50-node test instances. We systematically modify key components to assess their impact. First, we analyze the effects of different noise schedules and iteration steps on performance, as shown in Figure \ref{fig:ablation_step}. Second, we examine the impact of key components using Table \ref{tab:ablation}. In the W/o DML (Dual-Modal Learning) variant, we substitute our proposed graph transformer with the anisotropic graph neural network introduced in \cite{Sun2023}. In the W/o EI (Efficient Iteration) variant, we replace our unique iteration strategy with the standard DDIM paradigm employed in \cite{Sun2023, ma2024, li2024}. Figure \ref{fig:ablation_step} illustrates that our inverse function noise schedule generally outperforms other schedules across various iteration steps. Table \ref{tab:ablation}, which compares these modifications, shows that our graph transformer network consistently achieves superior results in both average solution length and inference time, with the performance gap widening as the number of iteration steps increases. Additionally, the table indicates that our iteration process significantly surpasses the standard DDIM paradigm. Notably, our 1-step results outperform DDIM's 10-step results, and our 5-step results exceed DDIM's 50-step results, highlighting the efficiency and effectiveness of our approach.

\begin{table}[t]
\centering
\caption{Ablation studies on our proposed components}
\label{tab:ablation}
\resizebox{\columnwidth}{!}{
\begin{tabular}{lccccccccc}
\toprule
\multicolumn{2}{l}{\textbf{Iteration  Step}} & 1     & 5     & 10     & 15     & 20     & 30     & 40     & 50     \\
\midrule
\multirow{2}{*}{W/o DML}     & Gap(\%)   & 0.24  & 0.09  & 0.07   & 0.07   & 0.05   & 0.06   & 0.05   & 0.05   \\
                         & Time(s)   & 20.00 & 56.20 & 107.67 & 158.15 & 212.88 & 319.88 & 425.89 & 529.35 \\
\midrule
\multirow{2}{*}{W/o EI}    & Gap(\%)   & 1.68  & 0.23  & 0.11   & 0.10   & 0.09   & 0.10   & 0.11   & 0.11   \\
                         & Time(s)   & 13.64 & 32.88 & 62.99  & 119.25 & 123.83 & 184.33 & 253.63 & 305.93 \\
\midrule
\multirow{2}{*}{DEITSP}  & Gap(\%)   & 0.13  & 0.03  & 0.02   & 0.01   & 0.01   & 0.01   & 0.01   & 0.00   \\
                         & Time(s)   & 14.95 & 37.85 & 72.00  & 127.62 & 146.09 & 218.14 & 297.14 & 369.81 \\
\bottomrule

\end{tabular}}

\end{table}

\section{Conclusion}
In this paper, we introduced a diffusion-based non-autoregressive model named DEITSP, which is specifically designed for solving the Traveling Salesman Problem (TSP). We proposed a dual-modality graph transformer and an efficient iterative strategy, achieving superior solution quality and inference speed compared to state-of-the-art methods. DEITSP's flexibility in adjusting iteration steps allows it to balance computation time with solution accuracy, and its strong generalization capability is evident across diverse real-world instances and benchmarks in our experiments.

While current diffusion-based approaches demonstrate strong performance, training with labeled data can present challenges when scaling to extremely large instances. We recognize this as an area for further improvement and will explore more efficient methods to enhance scalability, such as decomposing problems into smaller sub-problems \cite{Fu2021} or designing new unsupervised training mechanisms. Additionally, future work will focus on extending DEITSP to address a broader range of optimization problems, incorporating more complex constraints and varied environments, and further enhancing its robustness and applicability.

\bibliographystyle{ACM-Reference-Format}
\bibliography{MyBib1}

\appendix

\section{Decoding Strategy}
\label{decoding}

The diffusion model generates an adjacency matrix heatmap score that represents the connection probability of each edge. However, these adjacency matrices cannot ensure the feasibility of meeting TSP constraints. Therefore, a specialized decoding strategy is required to produce feasible solutions. We employ a combination of Greedy decoding and 2-opt as the default strategy for our experiments, following \citet{Sun2023,graikos2022}. In the greedy decoding method, we sort the edges based on $(\tilde{a}_{ij} + \tilde{a}_{ji})/cost(v_i,v_j)$, and degressively insert them into the partial solution while ensuring no conflict.

\section{Noise Schedule}
\label{ap:schedule}

\begin{figure}[ht]
    \centering
    \includegraphics[width=0.36\textwidth]{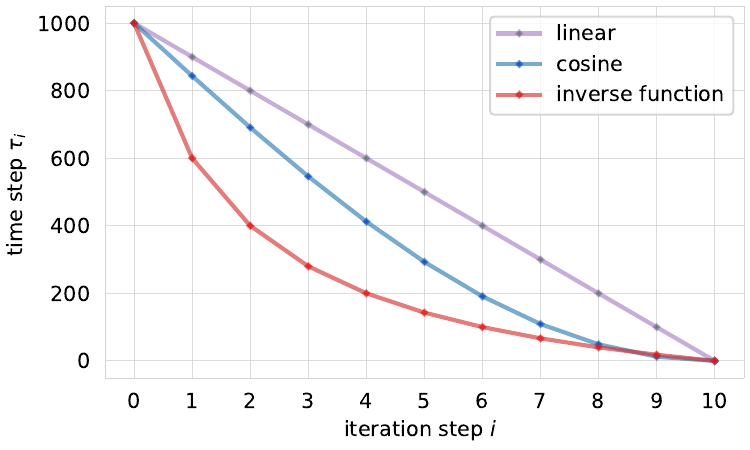}
    \caption{Comparison of noise step values with different iteration schedules (linear, cosine, and inverse function).}
    \Description{Comparison of noise step values with different iteration schedules}
    \label{fig:denoise_schedule}
\end{figure}

\begin{figure*}[ht]
    \centering
    \includegraphics[width=0.58\textwidth]{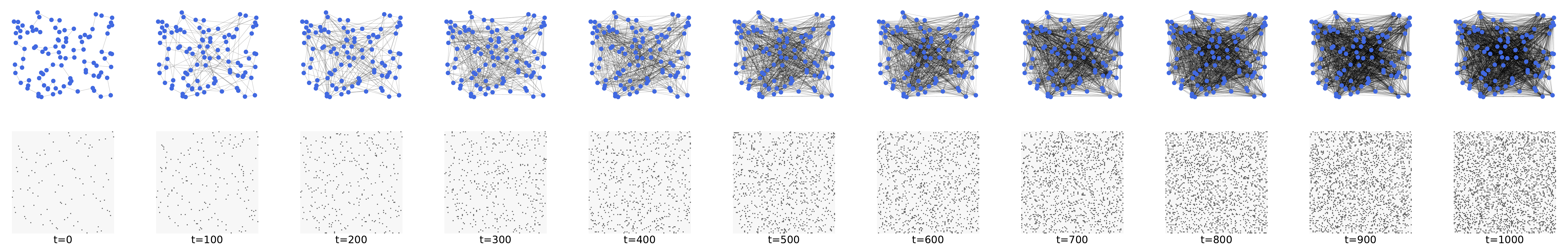}
    \caption{Visualization of noise addition process on the TSP100 instance.}
    \Description{Visualization of solutions produced by DEITSP using 4 step iterations.}
    \label{fig:addnoise_100}
\end{figure*}

As depicted in Figures \ref{fig:addnoise_20}, \ref{fig:addnoise_50}, and \ref{fig:addnoise_100}, the sparsity of TSP solutions leads to a significant difference in distribution between the optimal solution and the high-noise state. While higher noise levels expand the exploration range, they also increase inaccuracies in network predictions. Therefore, our proposed schedule, unlike the conventional linear and cosine schedules used in image generation, prioritizes more iterations in a low-noise state. To illustrate, Figure \ref{fig:denoise_schedule} presents the time step values for different schedules, using 10 steps as an example.

\section{Generalization to Real-World Instances}
\label{app:tsplib}

Unlike the uniformly sampled node coordinates used in training, the node coordinates in real-world TSPLIB instances are derived from authentic or carefully designed scenarios, reflecting structural and contextual similarities to real-world challenges. We apply models trained on TSP100 instances directly to 26 real-world instances, a strategy known as zero-shot generalization \cite{Joshi2022}. The ability of a model to generate satisfactory solutions under these conditions highlights its proficiency in recognizing and learning the fundamental patterns of TSPs, making it suitable for a wide range of applications beyond mere pattern recognition. Tables \ref{tab:tsplib_1} and Table \ref{tab:tsplib_2} show the performance of neural network-based approaches on 26 TSPLIB instances.

\section{Visualization of the Noise Addition Process}
\begin{figure*}[ht]
    \centering
    \includegraphics[width=0.58\textwidth]{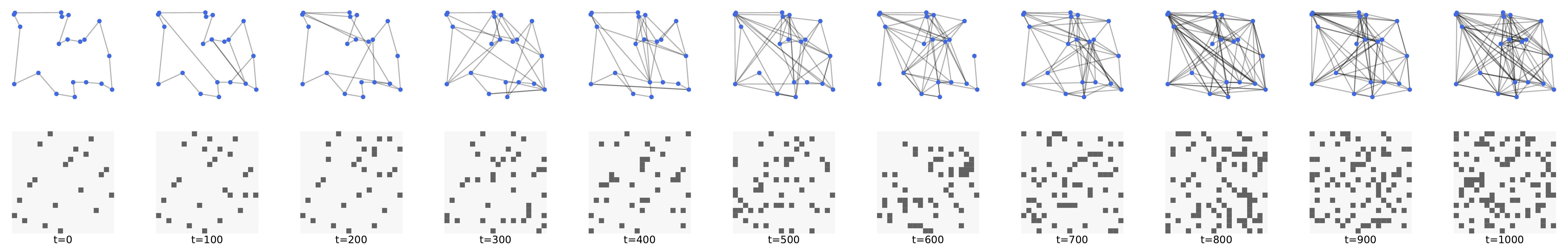}
    \caption{Visualization of noise addition process on the TSP20 instance.}
    \Description{Visualization of solutions produced by DEITSP using 4 step iterations.}
    \label{fig:addnoise_20}
\end{figure*}

\begin{figure*}[ht]
    \centering
    \includegraphics[width=0.58\textwidth]{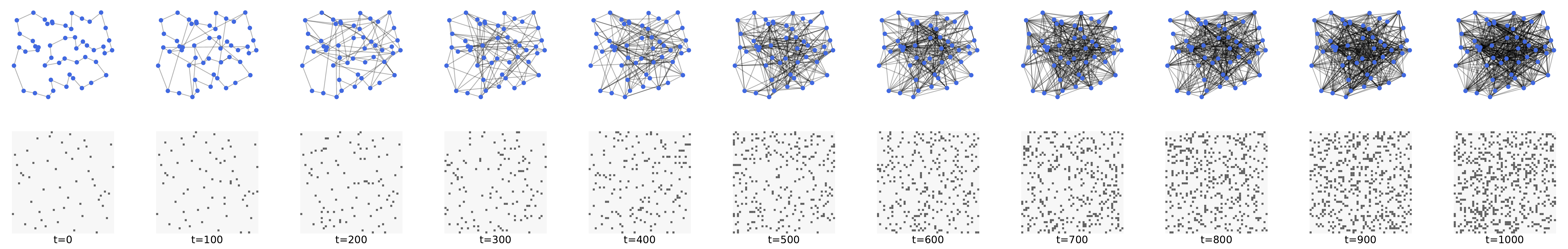}
    \caption{Visualization of noise addition process on the TSP50 instance.}
    \Description{Visualization of solutions produced by DEITSP using 4 step iterations.}
    \label{fig:addnoise_50}
\end{figure*}

We visualize the noise addition process in DEITSP, which disrupts the optimal TSP solution. We utilize instances of TSP20 (TSP with 20 nodes), TSP50, and TSP100, and set them according to the Eq. (\ref{addnoise}), where $t$ is the parameter representing the noise level. We visualize the noisy instances in the form of graphs and adjacency matrices. As shown in the Figure \ref{fig:addnoise_20} and Figure \ref{fig:addnoise_50}, with the increase of noise level, the adjacency matrix of TSP instances gradually transforms into a uniformly distributed pattern.

\section{Visualization of the Inference Process}

\begin{figure}[ht]
    \centering
    \includegraphics[width=0.45\textwidth]{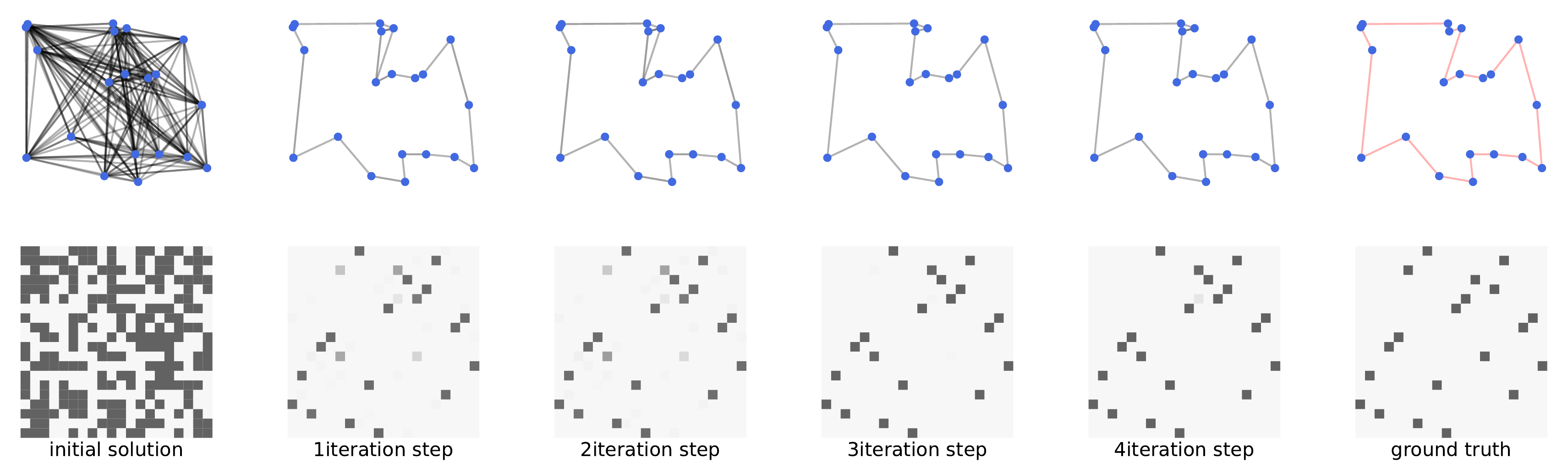}
    \caption{Visualization of solutions produced by DEITSP using 4 iteration steps on TSP20 instances.}
    \Description{Visualization of solutions produced by DEITSP using 4 step iterations.}
    \label{fig:visual_20}
\end{figure}

\begin{figure}[ht]
    \centering
    \includegraphics[width=0.45\textwidth]{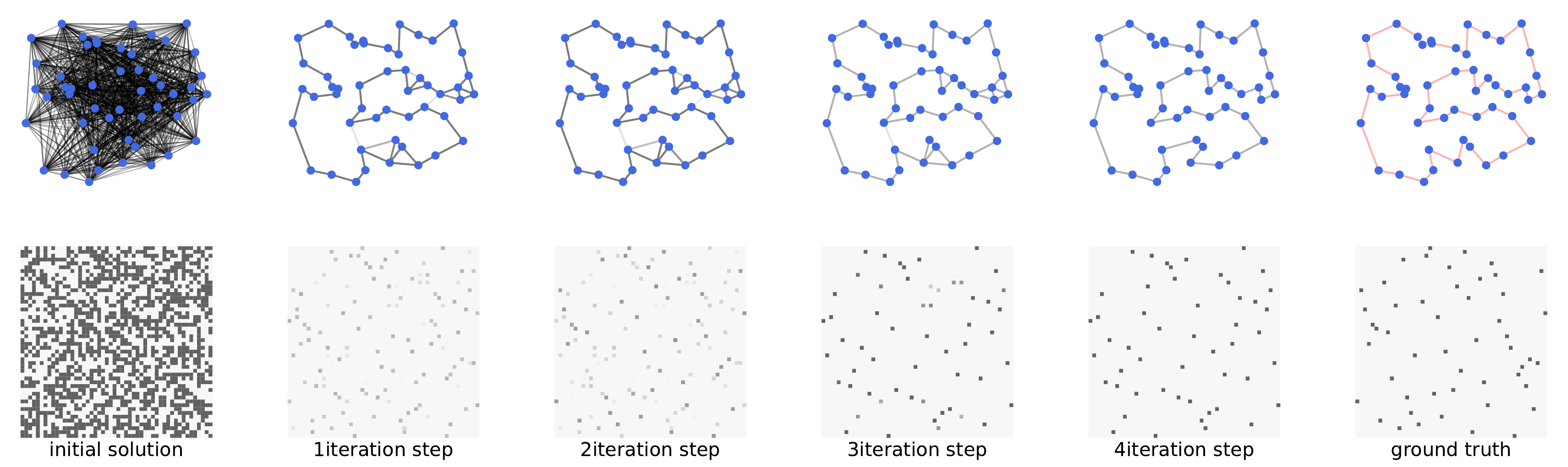}
    \caption{Visualization of solutions produced by DEITSP using 4 iteration steps on TSP50 instances.}
    \Description{Visualization of solutions produced by DEITSP using 4 step iterations.}
    \label{fig:visual_50}
\end{figure}

\begin{figure}[ht]
    \centering
    \includegraphics[width=0.45\textwidth]{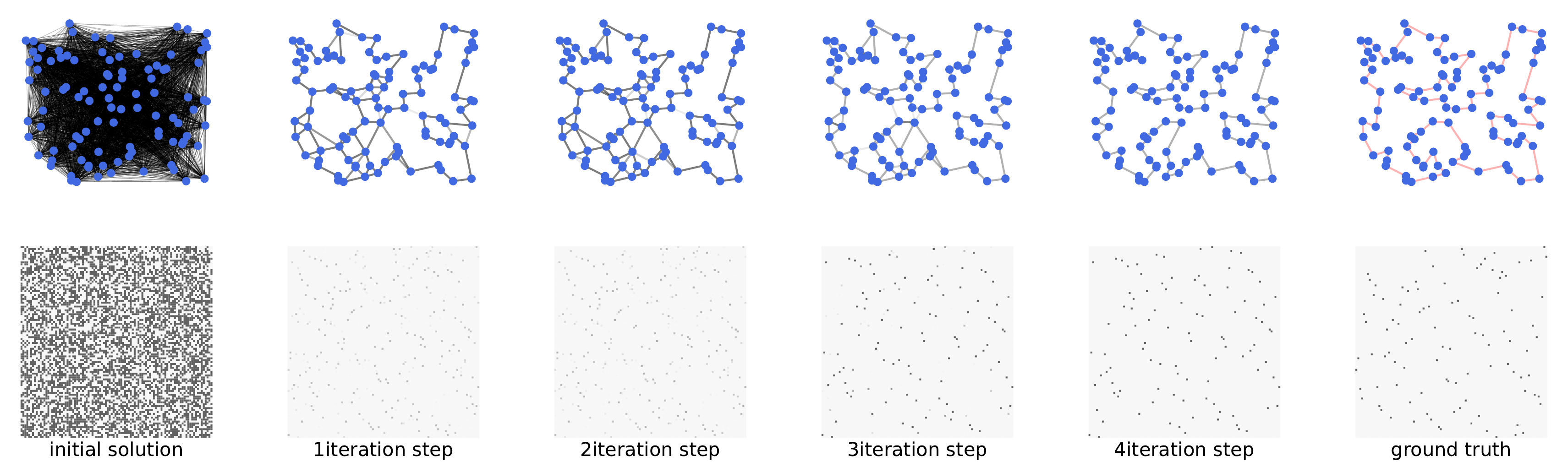}
    \caption{Visualization of solutions produced by DEITSP using 4 iteration steps on TSP100 instances.}
    \Description{Visualization of solutions produced by DEITSP using 4 step iterations.}
    \label{fig:visual_100}
\end{figure}

To visually illustrate the iterative exploration of multiple solutions during the DEITSP inference process, we present visualizations of the process. We utilize randomly generated instances of TSP20, TSP50 and TSP100, adhering to the same experimental settings outlined in Section \ref{hyperpara}. These visualizations encompass the initial solution derived from uniform sampling, the solution predicted by the diffusion model, and the ground truth, all presented in the form of graphs and adjacency matrices. As evident from Figures \ref{fig:visual_20}, \ref{fig:visual_50} and \ref{fig:visual_100}, DEITSP demonstrates its ability to predict high-quality solutions through a single-step denoising process and further enhances its performance by exploring a broader range of solutions over iterations. Notably, as the number of iteration steps increases, the quality of the predicted solutions gradually improves, thereby validating the efficacy of our noise scheduling strategy in optimizing the solution space.

\begin{table*}[t]
  \centering
  \caption{Generalization performance of applying models trained using TSP100 instances on TSPLIB instances (Part 1)}
  \label{tab:tsplib_1}
    \resizebox{0.70\textwidth}{!}{
  \begin{tabular}{lcccccccccccccccc}
  \toprule
  Name     & Optimal Len & \multicolumn{3}{c}{AM}           & \multicolumn{3}{c}{POMO}         & \multicolumn{3}{c}{Sym-NCO}      & \multicolumn{3}{c}{ELG}          & \multicolumn{3}{c}{HierTSP}      \\
            &                & Len        & Gap (\%) & Time (s) & Len        & Gap (\%) & Time (s) & Len        & Gap (\%) & Time (s) & Len        & Gap (\%) & Time (s) & Len        & Gap (\%) & Time (s) \\
  \midrule
  berlin52 & 7542.000       & 7856.426   & 4.169    & 0.129    & 7544.366   & 0.031    & 0.837    & 7544.662   & 0.035    & 0.851    & 7544.365   & 0.031    & 0.917    & 7544.662   & 0.035    & 0.872    \\
  bier127  & 118282.000     & 125270.101 & 5.908    & 0.522    & 123319.180 & 4.259    & 0.086    & 123993.359 & 4.829    & 0.092    & 122855.531 & 3.867    & 0.244    & 124093.656 & 4.913    & 0.125    \\
  ch130    & 6110.000       & 6304.420   & 3.182    & 0.550    & 6122.857   & 0.210    & 0.088    & 6118.715   & 0.143    & 0.090    & 6119.898   & 0.162    & 0.247    & 6157.908   & 0.784    & 0.131    \\
  ch150    & 6528.000       & 6827.244   & 4.584    & 0.697    & 6562.099   & 0.522    & 0.102    & 6567.454   & 0.604    & 0.104    & 6583.826   & 0.855    & 0.288    & 6578.063   & 0.767    & 0.147    \\
  eil101   & 629.000        & 647.832    & 2.994    & 0.359    & 640.596    & 1.844    & 0.070    & 640.212    & 1.782    & 0.071    & 643.840    & 2.359    & 0.193    & 642.830    & 2.199    & 0.100    \\
  eil51    & 426.000        & 432.935    & 1.628    & 0.125    & 431.953    & 1.397    & 0.037    & 431.953    & 1.397    & 0.038    & 430.746    & 1.114    & 0.098    & 429.484    & 0.818    & 0.054    \\
  eil76    & 538.000        & 548.717    & 1.992    & 0.225    & 544.369    & 1.184    & 0.053    & 544.652    & 1.236    & 0.055    & 549.433    & 2.125    & 0.146    & 547.020    & 1.677    & 0.078    \\
  kroA100  & 21282.000      & 22136.898  & 4.017    & 0.352    & 21396.438  & 0.538    & 0.071    & 21357.164  & 0.353    & 0.073    & 21307.422  & 0.119    & 0.192    & 21487.941  & 0.968    & 0.102    \\
  kroA150  & 26524.000      & 27527.668  & 3.784    & 0.695    & 26734.875  & 0.795    & 0.100    & 26890.738  & 1.383    & 0.103    & 26805.207  & 1.060    & 0.286    & 26995.578  & 1.778    & 0.146    \\
  kroA200  & 29368.000      & 31454.890  & 7.106    & 1.150    & 29984.789  & 2.100    & 0.134    & 30206.396  & 2.855    & 0.137    & 29831.221  & 1.577    & 0.383    & 30149.805  & 2.662    & 0.188    \\
  kroB100  & 22141.000      & 23279.490  & 5.142    & 0.352    & 22275.605  & 0.608    & 0.068    & 22374.285  & 1.054    & 0.070    & 22280.641  & 0.631    & 0.191    & 22275.350  & 0.607    & 0.099    \\
  kroB150  & 26130.000      & 26766.788  & 2.437    & 0.696    & 26635.195  & 1.933    & 0.100    & 26816.086  & 2.626    & 0.103    & 26374.766  & 0.937    & 0.288    & 26638.965  & 1.948    & 0.148    \\
  kroB200  & 29437.000      & 31951.214  & 8.541    & 1.150    & 30428.590  & 3.369    & 0.136    & 30563.463  & 3.827    & 0.140    & 29904.586  & 1.588    & 0.385    & 30551.854  & 3.787    & 0.188    \\
  kroC100  & 20749.000      & 20950.680  & 0.972    & 0.352    & 20832.773  & 0.404    & 0.068    & 20959.719  & 1.016    & 0.070    & 20770.041  & 0.101    & 0.194    & 20829.061  & 0.386    & 0.099    \\
  kroD100  & 21294.000      & 21872.558  & 2.717    & 0.352    & 21719.195  & 1.997    & 0.068    & 21635.557  & 1.604    & 0.070    & 21532.164  & 1.118    & 0.191    & 21772.988  & 2.249    & 0.102    \\
  kroE100  & 22068.000      & 22392.400  & 1.470    & 0.352    & 22380.355  & 1.415    & 0.068    & 22346.006  & 1.260    & 0.070    & 22237.137  & 0.766    & 0.194    & 22260.621  & 0.873    & 0.100    \\
  lin105   & 14379.000      & 14629.051  & 1.739    & 0.380    & 14494.633  & 0.804    & 0.072    & 14648.303  & 1.873    & 0.074    & 14467.038  & 0.612    & 0.200    & 14720.170  & 2.373    & 0.103    \\
  pr107    & 44303.000      & 46045.437  & 3.933    & 0.391    & 44897.246  & 1.341    & 0.073    & 45324.367  & 2.305    & 0.075    & 44960.422  & 1.484    & 0.205    & 44958.227  & 1.479    & 0.106    \\
  pr124    & 59030.000      & 61200.533  & 3.677    & 0.499    & 59091.836  & 0.105    & 0.083    & 59123.277  & 0.158    & 0.086    & 59181.652  & 0.257    & 0.236    & 59520.555  & 0.831    & 0.121    \\
  pr136    & 96772.000      & 101672.534 & 5.064    & 0.585    & 97485.609  & 0.737    & 0.095    & 97513.648  & 0.766    & 0.097    & 97733.406  & 0.993    & 0.259    & 98391.383  & 1.673    & 0.136    \\
  pr144    & 58537.000      & 63009.812  & 7.641    & 0.638    & 58828.539  & 0.498    & 0.097    & 59043.488  & 0.865    & 0.099    & 58859.398  & 0.551    & 0.275    & 58795.152  & 0.441    & 0.140    \\
  pr152    & 73682.000      & 79203.729  & 7.494    & 0.708    & 74440.711  & 1.030    & 0.102    & 76061.063  & 3.229    & 0.104    & 73720.609  & 0.052    & 0.292    & 74485.031  & 1.090    & 0.147    \\
  pr76     & 108159.000     & 109041.577 & 0.816    & 0.226    & 108159.438 & 0.000    & 0.053    & 108591.000 & 0.399    & 0.054    & 108444.047 & 0.264    & 0.145    & 108428.906 & 0.250    & 0.077    \\
  rat195   & 2323.000       & 2483.124   & 6.893    & 1.114    & 2558.470   & 10.136   & 0.130    & 2569.150   & 10.596   & 0.132    & 2400.775   & 3.348    & 0.372    & 2492.917   & 7.315    & 0.186    \\
  rat99    & 1211.000       & 1243.031   & 2.645    & 0.347    & 1273.635   & 5.172    & 0.068    & 1264.701   & 4.434    & 0.072    & 1248.142   & 3.067    & 0.189    & 1240.427   & 2.430    & 0.098    \\
  st70     & 675.000        & 686.725    & 1.737    & 0.200    & 677.110    & 0.313    & 0.052    & 677.110    & 0.313    & 0.051    & 677.110    & 0.313    & 0.134    & 677.110    & 0.313    & 0.071    \\
  \midrule
  Mean     & 31466.115      & 32703.968  & 3.934    & 0.506    & 31902.325  & 1.644    & 0.112    & 32069.482  & 1.959    & 0.115    & 31825.516  & 1.129    & 0.259    & 32025.602  & 1.717    & 0.149   \\
  \bottomrule
  \multicolumn{17}{l}{\large "Gap" indicate length differences compared to the optimal solutions from TSPLIB.}  
\end{tabular}}
\end{table*}

\begin{table*}[t]
  \centering
  \caption{Generalization performance of applying models trained using TSP100 instances on TSPLIB instances (Part 2)}
  \label{tab:tsplib_2}
  \resizebox{0.70\textwidth}{!}{
  \begin{tabular}{lcccccccccccccccc}
  \toprule
  Name     & Optimal Len & \multicolumn{3}{c}{DIFUSCO}      & \multicolumn{3}{c}{T2TCO}        & \multicolumn{3}{c}{\textbf{DEITSP}(1 Iter)} & \multicolumn{3}{c}{\textbf{DEITSP}(4 Iter)} & \multicolumn{3}{c}{\textbf{DEITSP}(16 Iter)} \\
            &                & Len        & Gap (\%) & Time (s) & Len        & Gap (\%) & Time (s) & Len         & Gap (\%)  & Time (s) & Len         & Gap (\%)  & Time (s) & Len           & Gap      & Time     \\
  \midrule
  berlin52 & 7542.000       & 7544.366   & 0.031    & 0.447    & 7544.366   & 0.031    & 0.749    & 7544.366    & 0.031     & 0.017    & 7544.366    & 0.031     & 0.047    & 7544.366      & 0.031    & 0.119    \\
  bier127  & 118282.000     & 119512.524 & 1.040    & 0.638    & 119050.160 & 0.649    & 0.953    & 119951.880  & 1.412     & 0.039    & 118885.507  & 0.510     & 0.096    & 119367.517    & 0.918    & 0.236    \\
  ch130    & 6110.000       & 6190.983   & 1.325    & 0.700    & 6122.493   & 0.204    & 1.035    & 6152.278    & 0.692     & 0.048    & 6152.278    & 0.692     & 0.113    & 6126.887      & 0.276    & 0.237    \\
  ch150    & 6528.000       & 6585.461   & 0.880    & 0.867    & 6580.991   & 0.812    & 1.295    & 6590.877    & 0.963     & 0.045    & 6580.099    & 0.798     & 0.106    & 6564.298      & 0.556    & 0.286    \\
  eil101   & 629.000        & 641.458    & 1.981    & 0.473    & 642.556    & 2.155    & 0.797    & 652.587     & 3.750     & 0.028    & 644.122     & 2.404     & 0.058    & 644.122       & 2.404    & 0.159    \\
  eil51    & 426.000        & 431.271    & 1.237    & 3.061    & 429.484    & 0.818    & 2.413    & 433.943     & 1.865     & 2.646    & 429.484     & 0.818     & 2.678    & 433.443       & 1.747    & 2.653    \\
  eil76    & 538.000        & 545.048    & 1.310    & 0.469    & 556.769    & 3.489    & 0.769    & 544.369     & 1.184     & 0.019    & 544.369     & 1.184     & 0.046    & 544.369       & 1.184    & 0.134    \\
  kroA100  & 21282.000      & 21285.443  & 0.016    & 0.483    & 21285.443  & 0.016    & 0.763    & 21307.422   & 0.119     & 0.031    & 21307.422   & 0.119     & 0.058    & 21307.422     & 0.119    & 0.159    \\
  kroA150  & 26524.000      & 26578.099  & 0.204    & 0.845    & 26525.031  & 0.004    & 1.251    & 27048.654   & 1.978     & 0.048    & 26900.181   & 1.418     & 0.103    & 26804.307     & 1.057    & 0.285    \\
  kroA200  & 29368.000      & 29583.978  & 0.735    & 1.414    & 30033.710  & 2.267    & 2.103    & 29718.696   & 1.194     & 0.072    & 29666.192   & 1.015     & 0.147    & 29543.848     & 0.599    & 0.490    \\
  kroB100  & 22141.000      & 22533.086  & 1.771    & 0.510    & 22645.401  & 2.278    & 0.772    & 22257.483   & 0.526     & 0.026    & 22303.429   & 0.734     & 0.055    & 22268.685     & 0.577    & 0.187    \\
  kroB150  & 26130.000      & 26284.696  & 0.592    & 0.850    & 26244.185  & 0.437    & 1.249    & 26547.005   & 1.596     & 0.046    & 26410.273   & 1.073     & 0.105    & 26319.820     & 0.726    & 0.289    \\
  kroB200  & 29437.000      & 30204.181  & 2.606    & 1.474    & 29841.325  & 1.374    & 2.102    & 30841.207   & 4.770     & 0.072    & 29898.069   & 1.566     & 0.171    & 29639.831     & 0.689    & 0.472    \\
  kroC100  & 20749.000      & 20773.073  & 0.116    & 0.492    & 20750.763  & 0.008    & 0.755    & 21299.404   & 2.653     & 0.027    & 20943.158   & 0.936     & 0.056    & 21293.943     & 2.626    & 0.170    \\
  kroD100  & 21294.000      & 21320.974  & 0.127    & 0.504    & 21294.291  & 0.001    & 0.778    & 21460.117   & 0.780     & 0.027    & 21460.117   & 0.780     & 0.064    & 21375.452     & 0.383    & 0.205    \\
  kroE100  & 22068.000      & 22372.583  & 1.380    & 0.498    & 22362.472  & 1.334    & 0.769    & 22806.209   & 3.345     & 0.024    & 22444.825   & 1.708     & 0.059    & 22427.985     & 1.631    & 0.168    \\
  lin105   & 14379.000      & 14432.139  & 0.370    & 0.524    & 14382.996  & 0.028    & 0.802    & 14382.996   & 0.028     & 0.037    & 14439.703   & 0.422     & 0.086    & 14382.996     & 0.028    & 0.183    \\
  pr107    & 44303.000      & 45551.263  & 2.818    & 0.531    & 44590.338  & 0.649    & 0.806    & 44897.401   & 1.342     & 0.032    & 44382.709   & 0.180     & 0.089    & 44519.916     & 0.490    & 0.179    \\
  pr124    & 59030.000      & 59814.535  & 1.329    & 0.602    & 59774.850  & 1.262    & 0.903    & 59632.923   & 1.021     & 0.032    & 59396.847   & 0.621     & 0.074    & 59607.740     & 0.979    & 0.234    \\
  pr136    & 96772.000      & 97664.229  & 0.922    & 0.711    & 96925.555  & 0.159    & 1.072    & 100361.346  & 3.709     & 0.048    & 98778.220   & 2.073     & 0.114    & 98097.238     & 1.369    & 0.252    \\
  pr144    & 58537.000      & 58853.339  & 0.540    & 0.767    & 59287.317  & 1.282    & 1.146    & 58620.693   & 0.143     & 0.059    & 58620.693   & 0.143     & 0.097    & 58604.905     & 0.116    & 0.272    \\
  pr152    & 73682.000      & 75613.977  & 2.622    & 0.865    & 74610.427  & 1.260    & 1.282    & 74448.987   & 1.041     & 0.068    & 75151.014   & 1.994     & 0.106    & 73683.641     & 0.002    & 0.292    \\
  pr76     & 108159.000     & 110019.494 & 1.720    & 0.439    & 109770.304 & 1.490    & 0.756    & 110247.698  & 1.931     & 0.016    & 109086.647  & 0.858     & 0.063    & 109160.652    & 0.926    & 0.131    \\
  rat195   & 2323.000       & 2392.573   & 2.995    & 1.371    & 2365.187   & 1.816    & 2.079    & 2366.034    & 1.853     & 0.058    & 2355.270    & 1.389     & 0.143    & 2350.777      & 1.196    & 0.452    \\
  rat99    & 1211.000       & 1220.098   & 0.751    & 0.562    & 1219.244   & 0.681    & 0.804    & 1221.318    & 0.852     & 0.033    & 1221.318    & 0.852     & 0.052    & 1219.244      & 0.681    & 0.153    \\
  st70     & 675.000        & 677.642    & 0.391    & 0.459    & 677.194    & 0.325    & 0.758    & 677.194     & 0.325     & 0.017    & 677.194     & 0.325     & 0.039    & 677.110       & 0.313    & 0.121    \\
  \midrule
  Mean     & 31466.115      & 31870.251  & 1.147    & 0.791    & 31750.494  & 0.955    & 1.114    & 32000.503   & 1.504     & 0.139    & 31777.827   & 0.948     & 0.185    & 31711.943     & 0.832    & 0.328   \\
  \bottomrule
  \multicolumn{17}{l}{\large "Gap" indicate length differences compared to the optimal solutions from TSPLIB.}  
\end{tabular}}
\end{table*}

\end{document}